\documentclass[runningheads]{llncs}

\usepackage{eccv}

\usepackage{eccvabbrv}

\usepackage{graphicx}
\usepackage{booktabs}
\usepackage{pdfpages}

\usepackage[accsupp]{axessibility}

\usepackage{lipsum}

\usepackage[utf8]{inputenc} 
\usepackage[T1]{fontenc}    
\usepackage{url}            
\usepackage{amsfonts}       
\usepackage{nicefrac}       
\usepackage{microtype}      
\usepackage{xcolor}         

\usepackage{threeparttable}
\usepackage{multirow}
\usepackage{enumitem}
\usepackage{graphicx}
\usepackage{blindtext}
\definecolor{mygray}{gray}{.9}
\makeatletter
\newcommand{\thickhline}{%
    \noalign {\ifnum 0=`}\fi \hrule height 1pt
    \futurelet \reserved@a \@xhline
}
\makeatother

\usepackage{array}

\newcommand{\ul}{\underline}

\usepackage{enumitem} 
\usepackage{subcaption} 
\usepackage{adjustbox} 
\usepackage{multirow} 
\usepackage{amsmath}
\usepackage{bbm} 
\usepackage{mathtools} 
\usepackage{wrapfig} 

\usepackage{colortbl} 

\usepackage{pifont}

\usepackage{tabularx}
\usepackage{float}

\newcolumntype{L}[1]{>{\raggedright\let\newline\\\arraybackslash\hspace{0pt}}m{#1}}
\newcolumntype{C}[1]{>{\centering\let\newline\\\arraybackslash\hspace{0pt}}m{#1}}

\usepackage[pagebackref,breaklinks,colorlinks,citecolor=eccvblue]{hyperref}

\usepackage{hyperref}

\usepackage{orcidlink}

\begin{document}

\title{Learning Probabilistic Prompt \\ for Continual Learning}

\author{Hyekang Park\inst{1}\orcidlink{0009-0006-5408-546X} \and
Sanghoon Lee\inst{1}\orcidlink{0009-0004-6380-1507} \and
Geon Lee\inst{1}\orcidlink{0000-0003-4217-9291} \and \\
Jongyoun Noh\inst{2}\orcidlink{0009-0004-9084-6524} \and
Bumsub Ham\inst{1}\thanks{Corresponding author.}\orcidlink{0000-0002-3443-8161} \\
\url{http://cvlab.yonsei.ac.kr/projects/ProbPrompt}
} 
\authorrunning{H.~Park et al.}

\institute{$^1$Yonsei University, Seoul, South Korea, $^2$Samsung Electronics, Suwon, South Korea}
\maketitle

\vspace{-3mm}
\begin{abstract}
Continual learning aims to progressively learn from a sequence of tasks, each containing a disjoint subset of classes, while preserving previously learned knowledge. Prompt-based continual learning methods propose to learn a small set of parameters,~\emph{i.e.},~prompts, by associating them with a query feature of an input image. These methods optimize the prompts, attempting to represent diverse patterns of images. However, we have observed that existing prompt-based methods suffer from a \emph{prompt collapse} problem, that is, the prompts tend to be highly similar to each other, thereby failing to capture the diverse data distributions in continual learning scenarios. To address this issue, we propose in this paper a novel prompt-based continual learning framework that captures diverse patterns of images across a sequence of tasks. To this end, we model each prompt as a probabilistic distribution and construct a mixture of these distributions, from which we sample diverse prompts. This enables our model to effectively capture highly diverse image distributions in the continual learning process. We also present a distribution regularization loss to prevent abrupt changes in the prompt distributions throughout the training process. We show extensive experimental results for continual learning on standard benchmarks, including ImageNet-R, CIFAR-100, and CUB-200, demonstrating the effectiveness of our framework.
\vspace{-2mm}
\end{abstract}

\vspace{-4mm}
\section{Introduction}
\label{sec:intro}

Deep neural networks often forget previously learned knowledge when continually learning new tasks~\cite{li2017learning}. This is mainly because the network parameters are overwritten during the continual learning process, which in turn interferes with preserving prior knowledge of a network. To address this problem, conventional approaches adopt regularization techniques~\cite{aljundi2018memory,ebrahimi2019uncertainty,kirkpatrick2017overcoming,titsias2019functional} or employ external memory to store previous data~\cite{aljundi2019online,aljundi2019gradient,chaudhry2018efficient}. These methods, however, are often suboptimal for large-scale models,~\eg,~vision transformers~(ViTs)~\cite{dosovitskiy2020image}, for which the forgetting problem becomes more severe~\cite{wang2022continual}. As an alternative, prompt-based continual learning methods have recently been proposed~\cite{wang2022dualprompt,wang2022learning,smith2023coda,jiao2024vector}. By introducing a small set of learnable prompts and optimizing the prompts while freezing the backbone, these methods encourage a network to effectively learn new tasks while retaining previously learned knowledge.

\begin{figure}[t]
    \centering
    \vspace{3mm}
    \begin{subfigure}{0.31\textwidth}
        \centering
        \includegraphics[width=\textwidth]{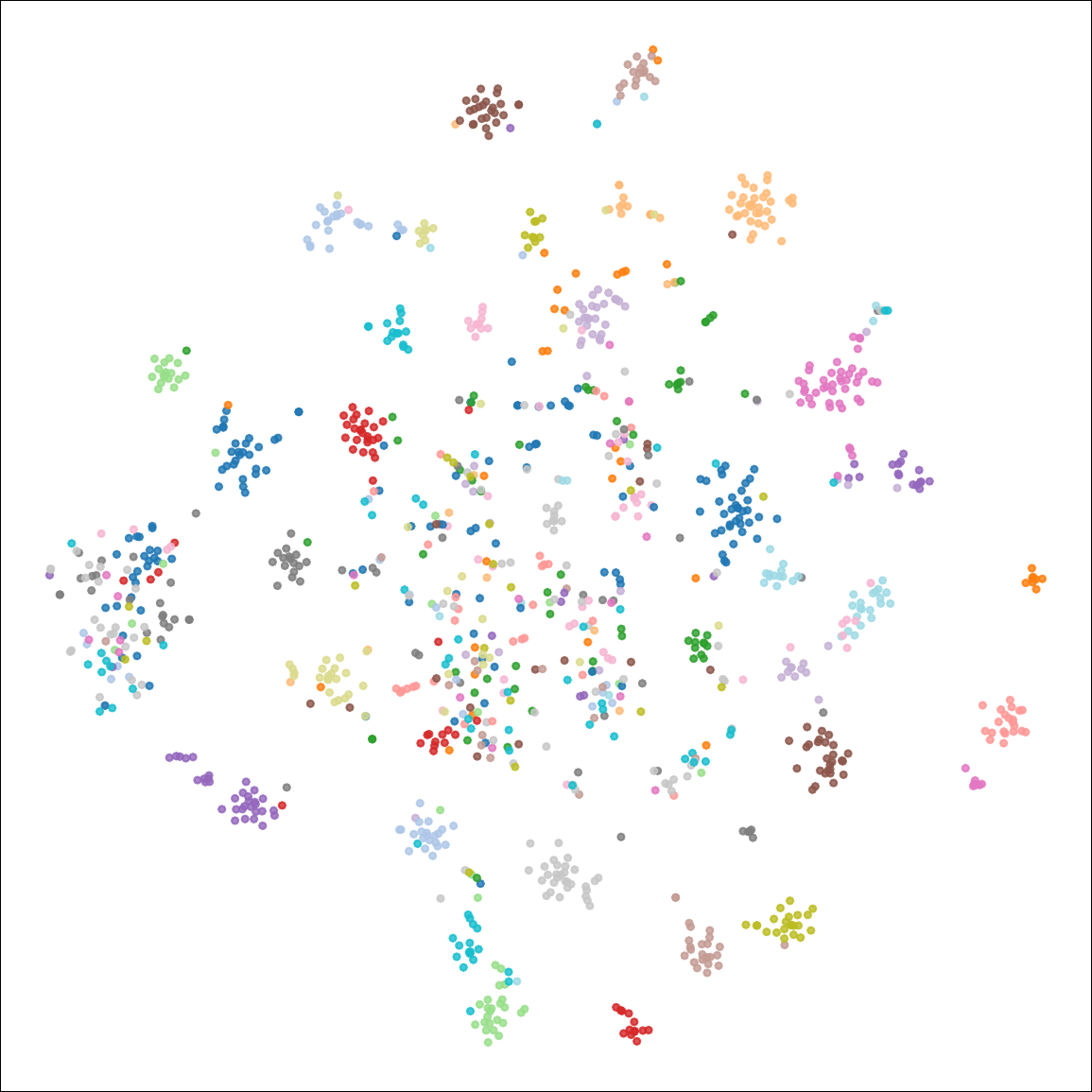}
        \caption{t-SNE visualization.}
        \label{fig:teaser_tsne}
    \end{subfigure}
    \hfill
    \begin{subfigure}{0.31\textwidth}
        \centering
        \includegraphics[width=\textwidth]{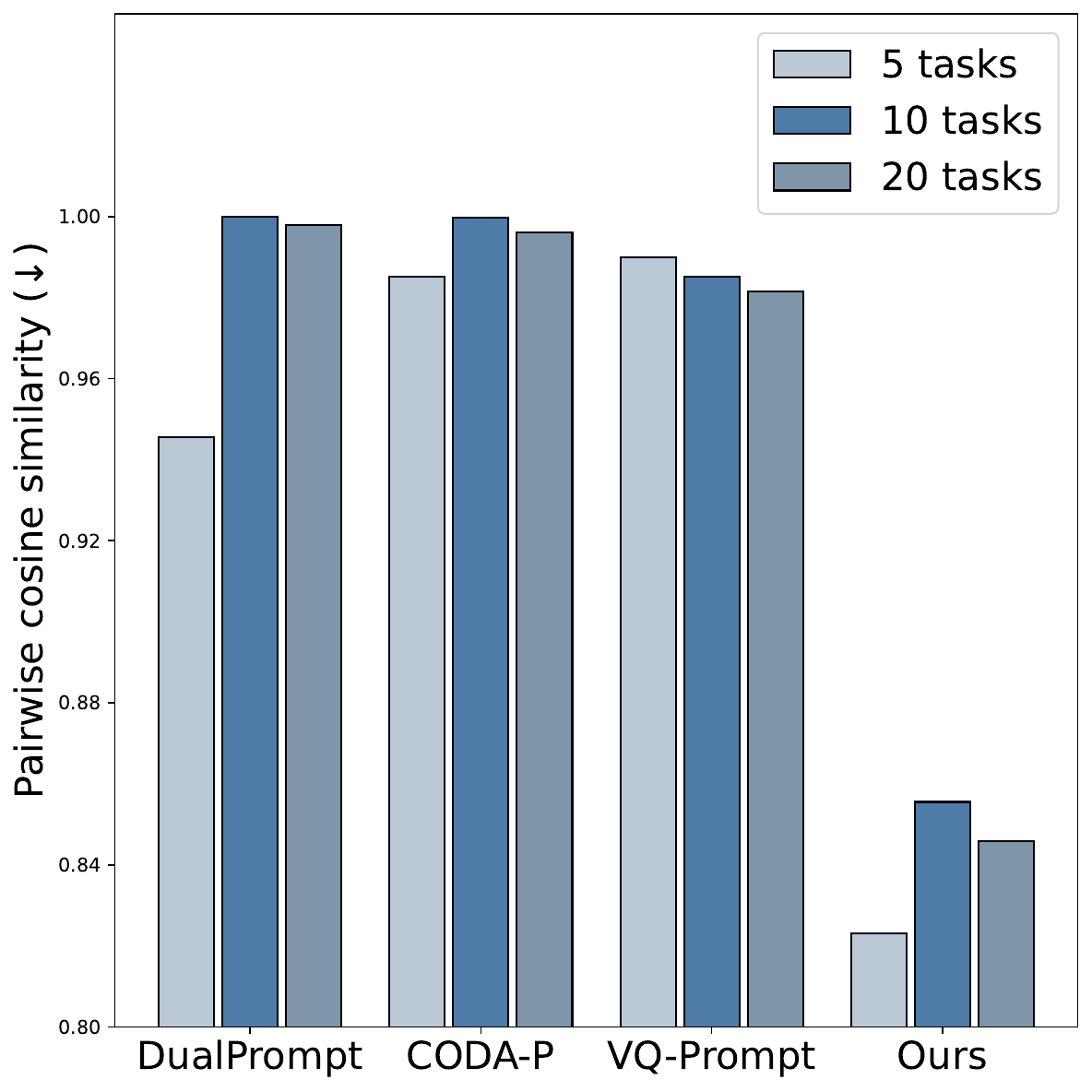}
        \caption{Pairwise prompt similarity.}
        \label{fig:teaser_cossim}
    \end{subfigure}
   \hfill
   \begin{subfigure}{0.31\textwidth}
       \centering
       \includegraphics[width=\textwidth]{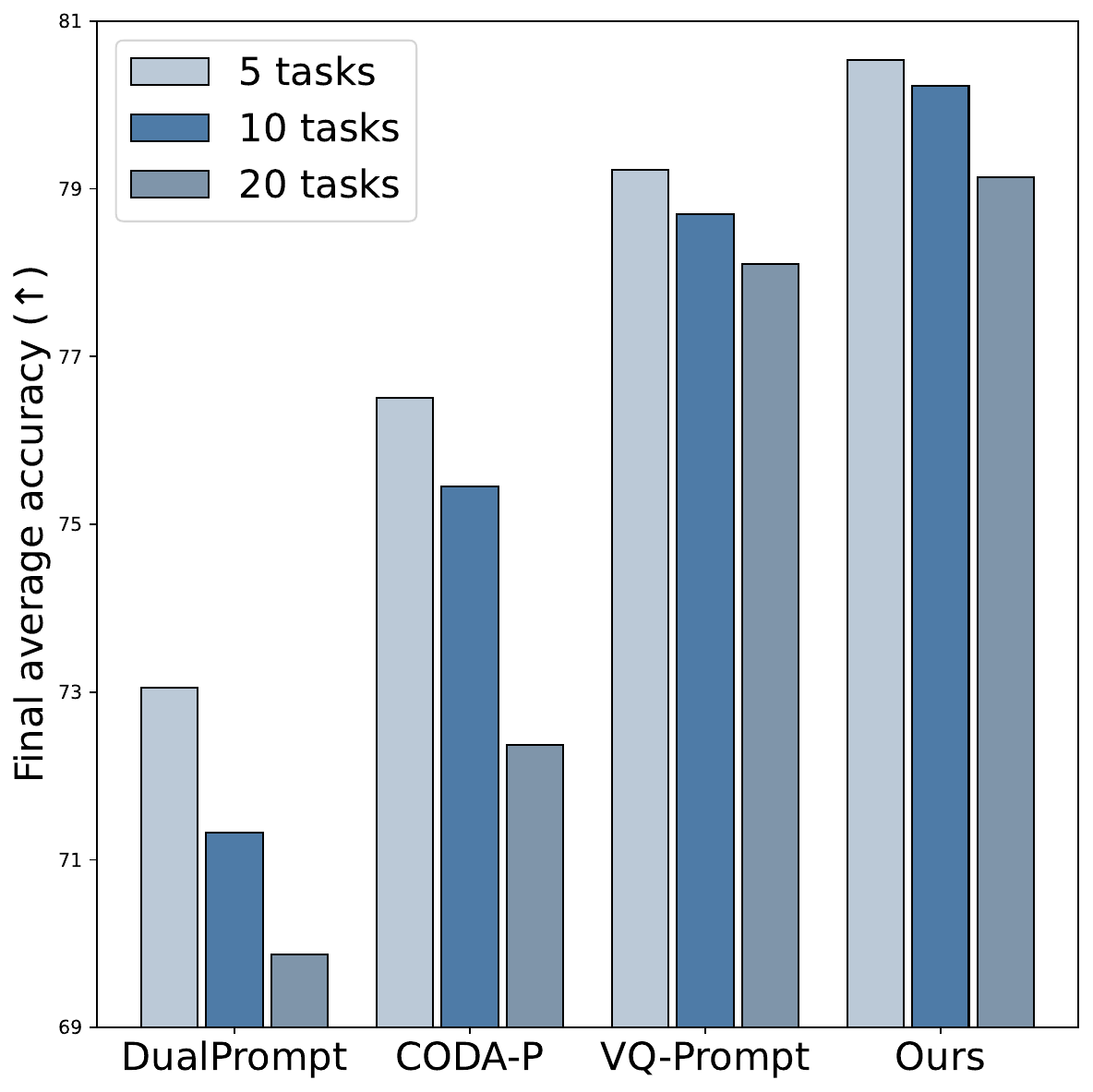}
       \caption{Final average accuracy.}
       \label{fig:teaser_faa}
   \end{subfigure}

    \vspace{-2mm}
    \caption{\textbf{(a)} t-SNE visualization of query features for the first task for 5 tasks on ImageNet-R~\cite{hendrycks2021many} (classes are color-coded). Note that the features are highly diverse even within the same incremental step. \textbf{(b)} Average cosine similarity between all pairs of prompts. Notably, our method achieves the highest diversity among prompts compared to existing works~\cite{wang2022dualprompt, jiao2024vector, smith2023coda}. \textbf{(c)} Quantitative comparison of final average accuracy across 5, 10, and 20 tasks on ImageNet-R~\cite{hendrycks2021many}.}
    \label{fig:teaser}
    \vspace{-8mm}
\end{figure}

Current prompt-based approaches~\cite{wang2022dualprompt,wang2022learning,smith2023coda,jiao2024vector} introduce an additional prompt token into a ViT encoder by leveraging a set of key-value pairs storing learnable prompts. Specifically, given a query feature representing an input image, these methods first associate the keys with the query. The matched values~(\ie, prompts) are then aggregated using, \eg, nearest neighbor selection~\cite{wang2022learning} or weighted summation~\cite{smith2023coda}, to represent the optimal prompt token for each image. This suggests that the efficacy of prompt-based approaches relies on how well the prompts can represent the distribution of the query features extracted from input images. In continual learning scenarios, however, a single task inherently exhibits high intra-task variations due to the presence of multiple distinct object classes~(see Fig.~\ref{fig:teaser}(a)). The feature variations between images become even more pronounced as the continual learning process progresses and new tasks are incrementally introduced, making the distribution highly complex to capture. Moreover, all current methods learn a static embedding to obtain prompt tokens, that is, they predominantly exploit a set of deterministic prompt vectors. Because the fixed vectors are restricted to representing specific values, their limited expressive power fails to properly capture the highly diverse patterns present across the sequence of tasks.

In this paper, we first identify a \emph{prompt collapse} problem that is prevalent across existing prompt-based approaches~\cite{wang2022dualprompt,wang2022learning,smith2023coda,jiao2024vector}, in which different prompt tokens become highly correlated with one another and fail to capture the diverse data distributions in the continual learning process~(Fig.~\ref{fig:teaser}(b)). To address this issue, we propose a novel prompt-based framework exploiting a set of probabilistic prompts, in contrast to the deterministic ones used in previous works. By modeling prompts as distributions, our framework encourages the prompts to better represent diverse patterns of images while avoiding the prompt collapse problem. Consequently, our probabilistic formulation yields consistent performance advantages over existing methods~(Fig.~\ref{fig:teaser}(c)). We also introduce a regularization scheme that enforces smooth transitions in prompt distributions during training, ensuring a stable continual learning process and better mitigating the forgetting problem. Extensive experiments on standard benchmarks demonstrate that our framework consistently outperforms state-of-the-art approaches~\cite{wang2022dualprompt,wang2022learning,smith2023coda,jiao2024vector} in terms of final average accuracy~(FAA) and cumulative average accuracy~(CAA) with a negligible computational overhead. Our main contributions can be summarized as follows:

\begin{itemize}[leftmargin=0.2in, label=\textbullet]
	\item We identify a prompt collapse problem among prompt-based continual learning frameworks, in which the prompt tokens fail to represent the diverse patterns present in a sequence of data. To address this limitation, we present a novel continual learning framework exploiting probabilistic prompts, enabling the model to better capture diverse query feature distributions.
	\item We introduce a distribution regularization loss that prevents abrupt changes in probabilistic prompts and better retains previously learned knowledge, thereby further mitigating the forgetting problem.
	\item We set a new state of the art on standard benchmarks, including CIFAR100~\cite{krizhevsky2009learning}, ImageNet-R~\cite{hendrycks2021many}, and CUB200~\cite{wah2011caltech}, demonstrating the effectiveness and efficiency of our method with extensive experimental results and ablation studies.
\end{itemize}

\section{Related Work}

Continual learning aims to learn new knowledge progressively, while avoiding the forgetting problem~\cite{li2017learning}. It is generally divided into three settings: Task-incremental learning~(TIL), domain-incremental learning~(DIL), and class-incre-mental learning (CIL) \cite{belouadah2021comprehensive,van2019three,de2021continual,masana2022class}. In TIL, training and evaluation are performed on distinct tasks with explicit task identifiers~\cite{van2019three,oh2022alife,baek2022decomposed}, while DIL focuses on generalizing across various domains that share the same label space~\cite{van2019three,masana2022class}. On the other hand, CIL attempts to classify previously learned classes jointly, without relying on any task information~\cite{masana2022class,de2021continual,van2019three}. To this end, CIL methods should preserve prior knowledge, while learning new classes continually, which is more challenging than other settings~\cite{masana2022class,de2021continual,van2019three,wang2022dualprompt,wang2022learning,smith2023coda,jiao2024vector}. A key challenge in CIL is thus balancing the plasticity and rigidity of a model, which are the abilities to learn new concepts and preserve previously learned knowledge, respectively~\cite{mermillod2013stability}. CIL methods can further be categorized into three groups based on how they address this trade-off: architecture-based, regularization-based, and rehearsal-based approaches. Architecture-based methods expand~\cite{yoon2017lifelong} or adapt~\cite{serra2018overcoming} networks to avoid forgetting, while acquiring knowledge from a new task. Regularization-based approaches~\cite{aljundi2018memory,chaudhry2018efficient,zenke2017continual} encourage parameters, pertinent to previous tasks, to be preserved during training, alleviating the forgetting problem. Rehearsal-based methods introduce a replay buffer to store past examples~\cite{rebuffi2017icarl,castro2018end} or latent features~\cite{tian2023continuous} explicitly, generally outperforming the other two approaches.

\vspace{2mm}
\noindent
\textbf{Prompt-based continual learning. } 
With the remarkable success of large-scale models, such as ViTs~\cite{dosovitskiy2020image}, the works of~\cite{douillard2022dytox,wang2022continual,pelosin2022towards,ermis2022memory} explore adopting these models for continual learning. Specifically, they modify transformer architectures to retain previous knowledge~\cite{douillard2022dytox,wang2022continual} or regularize attention scores~\cite{pelosin2022towards} to preserve previously acquired information. Albeit effective, these approaches require optimizing parameters of the large-scale models, which is computationally demanding. To alleviate this problem, prompt-based continual learning methods~\cite{wang2022learning,wang2022dualprompt,smith2023coda,jiao2024vector,kurniawan2024evolving,tang2023prompt,chen2025achieving,jeon2024rep} have recently been introduced. They freeze pre-trained backbones to preserve their representation ability for alleviating the forgetting problem, and employ learnable prompts. Early works of~\cite{wang2022learning,wang2022dualprompt} present a task-conditioned prompting to capture task-relevant knowledge. In particular, they use a pool of learnable prompts, where each prompt is paired with a key. Given an input image, they first extract a query feature, and then select prompts based on a key-query matching process, encouraging the prompts to learn task-specific knowledge during training. However, these methods cannot optimize prompts end-to-end, as they select the prompts using the non-differentiable key-query matching process,~\emph{e.g.}, nearest neighbor selection. To overcome this problem, recent approaches~\cite{smith2023coda,jiao2024vector} instead propose to learn prompt components, and generate prompt tokens for each input image dynamically by aggregating the components based on key-query similarities, encouraging an end-to-end optimization. Moreover, CODA-P~\cite{smith2023coda} uses an element-wise attention mechanism to focus more on relevant queries in the aggregation process. VQ-Prompt~\cite{jiao2024vector} attempts to mimic the categorical perception strategy of the human brain, and proposes to use discrete component vectors. Another line of works~\cite{tang2023prompt,kurniawan2024evolving} generate prompts using features from intermediate layers with additional trainable modules. While these prompt-based methods are effective for continual learning, they rely on deterministic prompts, resulting in prompt collapse, which limits to represent diverse data distributions present in a sequence of data. In contrast to the prior works, we propose to sample prompts stochastically from learned distributions to better represent diverse query features across a task, alleviating the prompt collapse problem.

\noindent
\textbf{Probabilistic representation learning. } 
Probabilistic representations have been exploited in various tasks in computer vision, including retrieval~\cite{chun2021probabilistic}, classification~\cite{schonfeld2019generalized}, and segmentation~\cite{oh2018modeling}. Latent features are represented as probabilistic distributions, rather than deterministic ones, enabling more flexible modeling of uncertainty and semantic variations. Recently, the works of~\cite{lu2022prompt,kwon2023probabilistic} have explored probabilistic modeling for prompt learning. Specifically, ProDA~\cite{lu2022prompt} enhances CLIP~\cite{radford2021learning} by representing each classifier weight as a probabilistic embedding. It estimates a distribution over weights using multiple textual prompts per class, trying to better capture the intra-class diversity of input images. PPL~\cite{kwon2023probabilistic} models attributes of objects in images as probabilistic prompts for dense prediction tasks, such as instance segmentation, and samples multiple prompts to generate granular textual representations for capturing image details. These approaches are closely related to ours in that they use probabilistic prompts. In contrast to these works, our approach differs in the following aspects:~(1)~The prompts themselves in~\cite{lu2022prompt,kwon2023probabilistic} are modeled as learnable but deterministic vectors, with probabilistic distributions defined over their output representations. Such output-level modeling makes it difficult to address prompt-level collapse directly. In contrast, we parameterize the prompts themselves as probabilistic distributions and construct a query-conditioned mixture distribution from which we sample prompts, thereby addressing prompt collapse at the prompt level\footnotemark.~(2)~We do not exploit additional information to learn new knowledge incrementally, whereas the works of~\cite{lu2022prompt,kwon2023probabilistic} require a task identifier or prior knowledge for classes in the context of class-incremental learning. We instead propose a novel framework that selects prompts considering each input to capture the diversity of query features across the sequence of tasks.~(3)~We introduce a distribution regularization loss to avoid an abrupt change in prompts during training, which can occur when applying existing probabilistic methods~\cite{lu2022prompt,kwon2023probabilistic} to continual learning scenarios without considering the rigidity of a model. The regularization term helps to maintain the rigidity of the model, effectively mitigating the forgetting problem.~(4)~Our framework does not require additional trainable modules or networks as in~\cite{kwon2023probabilistic,lu2022prompt}, except for the prompts and classifiers, and keeps pre-trained backbones frozen. This leads to greater efficiency and makes it more suitable for continual learning.

\footnotetext{More detailed comparisons between prompt- and output-level modeling can be found in the supplementary material.}

\begin{figure*}[t]
    \vspace{3mm}
    \centering
    \begin{subfigure}{0.185\textwidth}
        \centering
        \includegraphics[width=\textwidth]{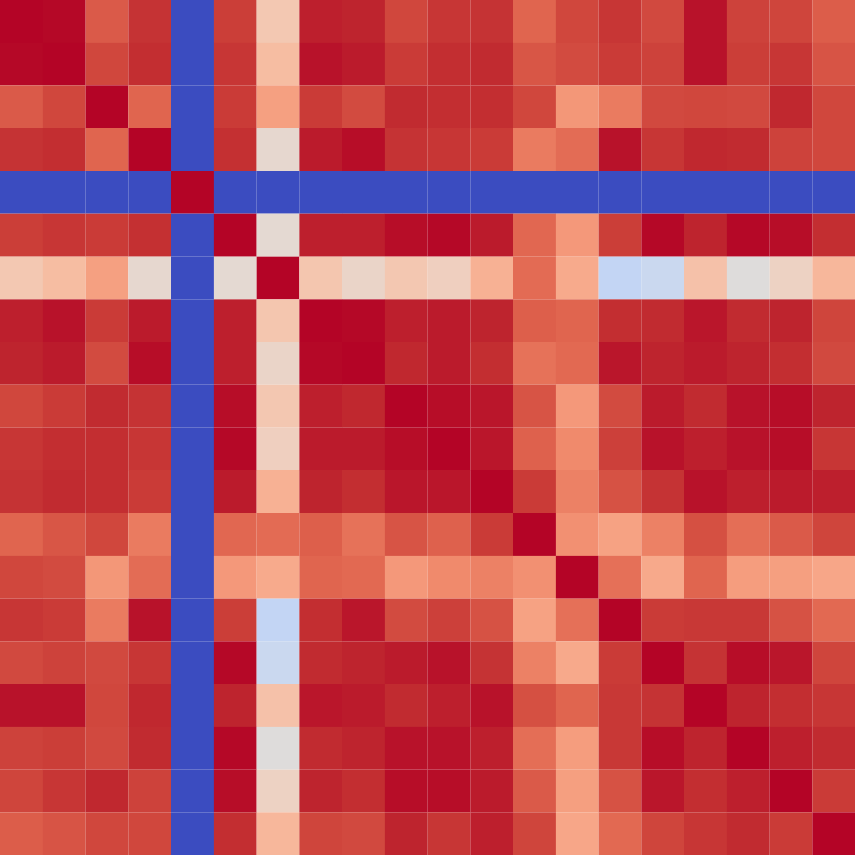}
        \caption{{L2P.}}
    \end{subfigure}
    \begin{subfigure}{0.185\textwidth}
        \centering
        \includegraphics[width=\textwidth]{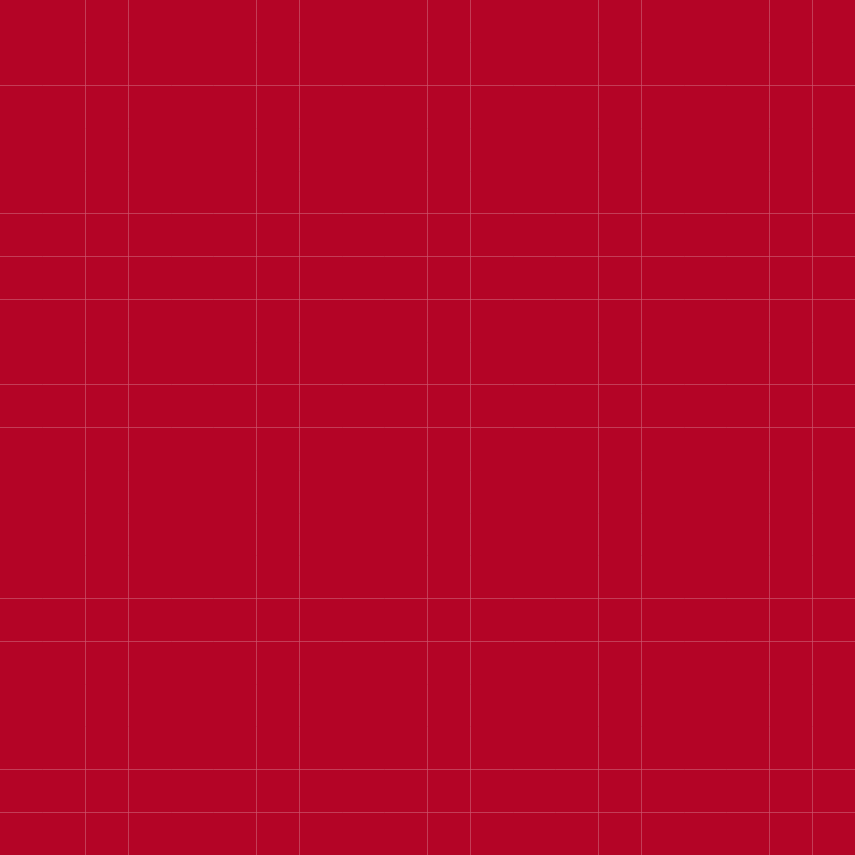}
        \caption{{DualPrompt.}}
    \end{subfigure}
    \begin{subfigure}{0.185\textwidth}
        \centering
        \includegraphics[width=\textwidth]{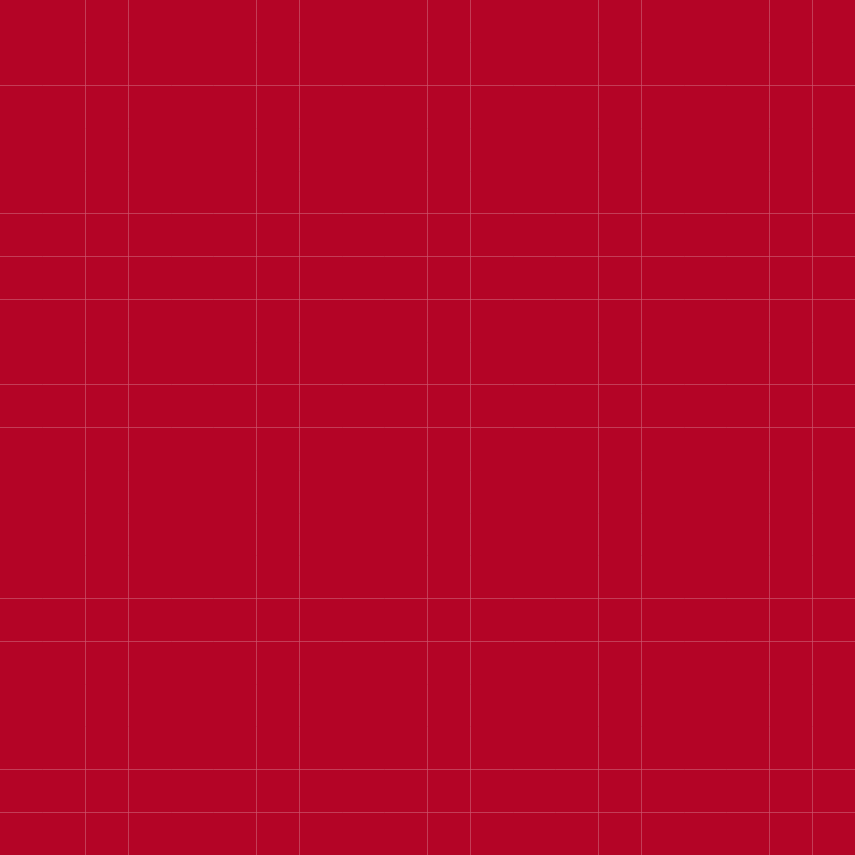}
        \caption{{VQ-Prompt.}}
    \end{subfigure}
    \begin{subfigure}{0.185\textwidth}
        \centering
        \includegraphics[width=\textwidth]{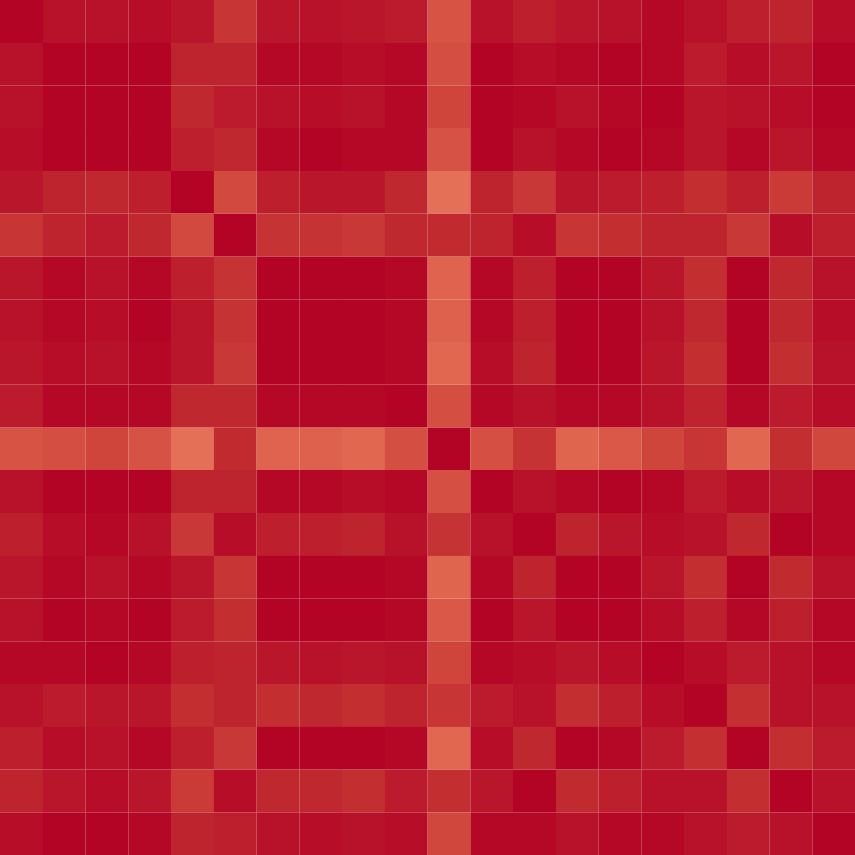}
        \caption{{CODA-P.}}
    \end{subfigure}
    \begin{subfigure}{0.185\textwidth}
        \centering
        \includegraphics[width=\textwidth]{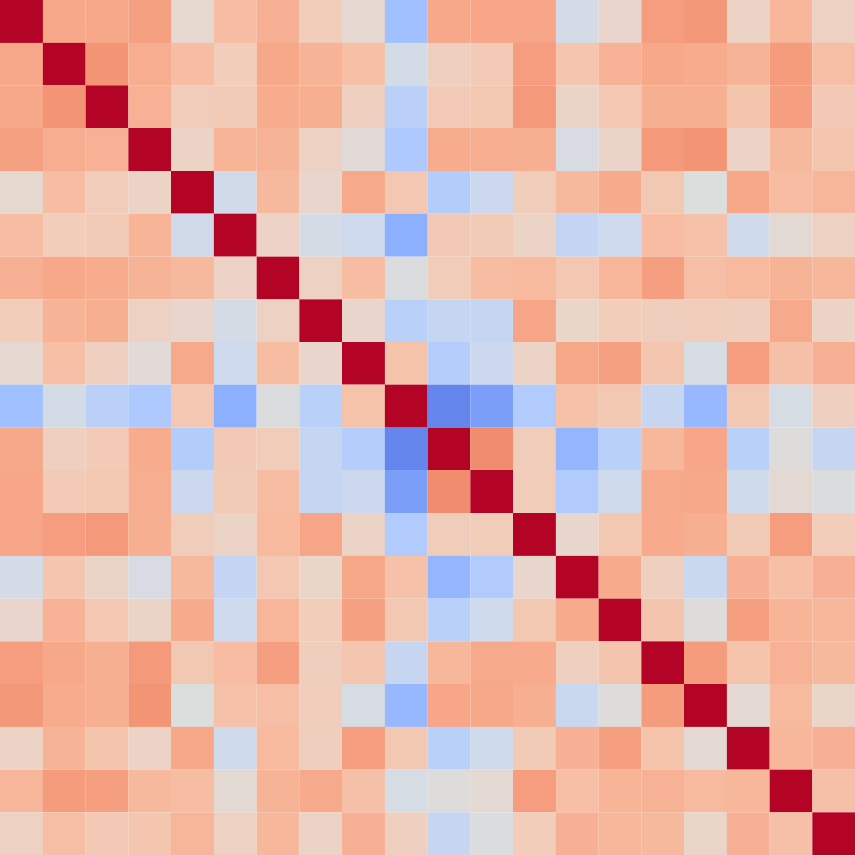}
        \caption{{{Ours.}}}
    \end{subfigure}
    \begin{minipage}[t]{0.029\textwidth}  
        \vspace{-19.7ex}  
        \includegraphics[width=\textwidth]{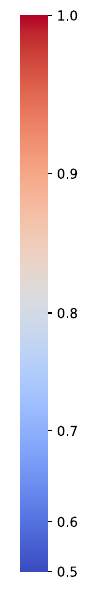}
    \end{minipage}
    \vspace{-6mm}
    \caption{ Visualization of cosine similarity between prompt tokens, obtained from \textbf{(a)}~L2P~\cite{wang2022learning}, \textbf{(b)}~DualPrompt~\cite{wang2022dualprompt}, \textbf{(c)}~VQ-Prompt~\cite{jiao2024vector}, \textbf{(d)} CODA-P~\cite{smith2023coda}, and \textbf{(e)} ours. The prompt tokens in \textbf{(a)}-\textbf{(e)} are obtained for the same query features, where each query feature is randomly selected from different classes for 10 tasks of ImageNet-R~\cite{hendrycks2021many}.}
    \label{fig:motivation}
    \vspace{-7mm}
\end{figure*}

\vspace{-3mm}
\section{Method}
\vspace{-2mm}

In this section, we first identify the prompt collapse problem based on the empirical observation~(Sec.~\ref{sec:3.1}) and describe our approach with an overview~(Sec.~\ref{sec:3.2}). We then present a probabilistic prompting framework in detail~(Sec.~\ref{sec:3.3}), including a training objective~(Sec.~\ref{sec:3.4}).

\vspace{-3mm}
\subsection{Prompt collapse problem} \label{sec:3.1} 
\vspace{-2mm}

Prompt-based CIL methods~\cite{wang2022learning,wang2022dualprompt,smith2023coda,jiao2024vector} optimize prompts without updating parameters in a backbone to preserve the representation capability of the pre-trained backbone, which is helpful for addressing the forgetting problem. Despite their effectiveness, we observe a critical limitation of existing prompt-based methods, the \emph{prompt collapse} problem, namely that different prompt tokens become highly similar to each other~(Fig.~\ref{fig:motivation}) and fail to represent the diverse data distributions across the sequence of tasks. Specifically, the feature variations between images become more pronounced as the continual learning process progresses, resulting in highly diverse data distributions. The existing methods fail to capture such complex distributions, since they set each prompt component as a fixed vector with limited representation capability. A straightforward way to mitigate this issue is to use more prompt components, but this still produces highly similar prompts, suggesting that prompt collapse cannot be alleviated by simply scaling up the number of prompt components\footnotemark. To address this, we propose a probabilistic prompt tuning framework that samples diverse prompts from learned distributions conditioned on query features.

\footnotetext{More detailed analyses and results can be found in the supplementary material.}

\subsection{Overview} \label{sec:3.2} 

\begin{figure*}[t]
    \vspace{3mm}
    \centering
    \includegraphics[width=\textwidth]{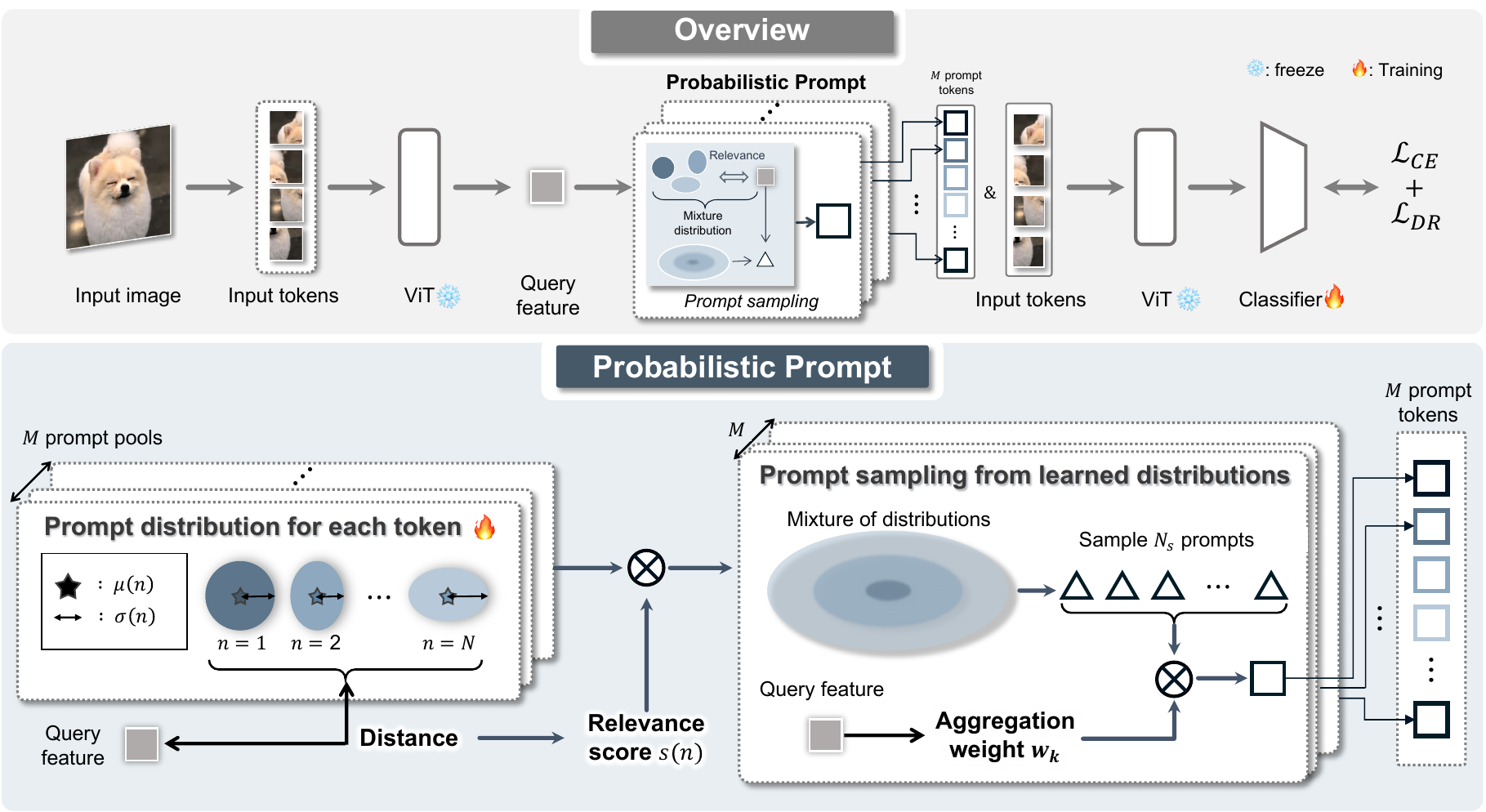}
    \vspace{-3mm}
    \caption{An overview of our framework. We form a mixture of prompt distributions by aggregating prompt distributions based on the distance between a query feature and each distribution, and sample prompt tokens from the mixture distribution. The sampled prompt tokens are prepended to input tokens to capture the diverse patterns of images in a sequence of data.}
    \label{fig:overview}
    \vspace{-6mm}
\end{figure*}

We introduce a novel prompt-based CIL framework that samples prompt tokens from probabilistic distributions to address the prompt collapse problem in previous methods. We show in Fig.~\ref{fig:overview} an overview of our framework. We set each prompt component as a Gaussian distribution, parameterized by mean and covariance, which we call a prompt distribution. Given a query feature from an input image, we compute the distance between the query and each prompt distribution. We then form a mixture of the prompt distributions, from which we sample diverse prompts to capture the various patterns of input data in the continual learning process. In particular, we sample multiple prompts and aggregate them to obtain the final prompt token while improving learning stability. This stochastic sampling encourages exploring diverse prompts to represent the various patterns of images~(Fig.~\ref{fig:motivation}(e)). Following the standard protocol in~\cite{wang2022dualprompt,smith2023coda,jiao2024vector}, we adopt a prefix tuning~(Pre-T) technique in order to provide representations of query features to a pre-trained model, by prompt tuning. Specifically, we prepend the sampled prompt tokens, concatenate the prompt tokens with input tokens, and input them into the pre-trained model. We optimize the prompt distributions end-to-end using a standard cross-entropy loss and our distribution regularization loss, over $T$ tasks continually. We do not update parameters in a backbone directly to preserve the representation ability of pre-trained backbones, which is helpful for retaining the previous knowledge~\cite{wang2022learning,wang2022dualprompt,smith2023coda,jiao2024vector,kurniawan2024evolving,tang2023prompt}.

\vspace{-3mm}
\subsection{Probabilistic prompt tuning} \label{sec:3.3} 

Deterministic pools of prompts or component vectors in existing prompt-based CIL methods suffer from a prompt collapse problem, which limits their ability to capture diverse patterns across task sequences. To overcome the prompt collapse problem, we introduce a probabilistic prompting framework that samples prompts from learned distributions. To this end, we model each pool of prompts as a set of probabilistic distributions, and sample prompt tokens from a mixture of the distributions, allowing the prompt tokens to consider diverse patterns of the queries. Following~\cite{wang2022learning,wang2022dualprompt,smith2023coda,jiao2024vector,kurniawan2024evolving,tang2023prompt}, we assume that each input is associated with $M$ prompt tokens, each of which is obtained from a corresponding pool of prompts. Each pool consists of $N$ prompt distributions. For the $m$-th pool, we assume that its $n$-th prompt is normally distributed as follows: 
\begin{equation} \label{eq:dist}
     \mathcal{N} \left( {\mu}(m, n), {\Sigma}(m, n) \right),
\end{equation}
where $m \in \{1, \dotsc, M\}$ and $n \in \{1, \dotsc, N \}$.
We denote by ${\mu}$ a $D$-dimensional mean vector, and ${\Sigma}$ a $D\times D$ diagonal covariance matrix with elements of variance, ${\sigma}^2(m, n) \in \mathbb{R}^D$, \ie,~${\Sigma}(m, n) = \mathrm{diag}({\sigma}^2(m, n))$. In the following, we omit the index of the prompt token $m$ for brevity, as the same process is applied to each pool independently.

Given a query feature ${q}$ from an input image, we form a mixture of the distributions to sample prompts for each prompt token. To be specific, we compute a relevance score~$s(n)$ for the $n$-th distributions of the pool w.r.t the query feature ${q}$ as follows:
 \begin{equation} \label{eq:score}
    s(n) = \frac{\exp\left\{-  S^2({q}, \mu(n), \Sigma(n))\right\}}{\sum_{n=1}^{N} \exp\left\{- S^2({q}, {\mu}(n), {\Sigma}(n))\right\}},
\end{equation}
where 
\begin{equation} \label{eq:maha_dist}
    S({q}, {\mu}(n), {\Sigma}(n)) = \sqrt{  \left({q} - {\mu}(n)\right)^{\top} \left({\Sigma}(n) \right)^{-1} \left({q} - {\mu}(n)\right)}.
\end{equation}
Note that ${\Sigma}(n)^{-1}$ is computed in an element-wise manner, since ${\Sigma}(n)$ is diagonal. We then compute a mixture distribution for each input token using the relevance scores. That is, the mixture distribution for the pool is defined as follows: 
\begin{equation} \label{eq:mixture}
    \mathcal{N}_{\mathrm{GM}} \left(  {\mu}_{\mathrm{GM}}, {\Sigma}_{\mathrm{GM}} \right),
 \end{equation}
where 
 \begin{equation} \label{eq:mix_mean}
     {\mu}_{\mathrm{GM}} = \sum_{n} s(n) {\mu}(n),
 \end{equation}
and 
\begin{equation} 
    {\Sigma}_{\mathrm{GM}} = \mathrm{diag}({\sigma}_{\mathrm{GM}}^2),	
\end{equation}
with 
\begin{equation} \label{eq:mix_var}
    {\sigma}_{\mathrm{GM}}^2 = \sum_{n} s(n) \left( {\sigma}^{2}(n) + \left( {\mu}(n) - {\mu}_{\mathrm{GM}} \right)^{2}\right).
\end{equation}

From the mixture distribution in Eq.~\eqref{eq:mixture}, we sample a probabilistic prompt $\tilde{P} $ for a prompt token using the reparameterization trick~\cite{kingma2015variational}, enabling training our model end-to-end:
\begin{equation} \label{eq:sampling}
    \tilde{P} =  {\mu}_{\mathrm{GM}} +  {\sigma}_{\mathrm{GM}} \odot  {\epsilon}, \text{~where~} {\epsilon} \sim \mathcal{N}\left( 0, \mathbf{I} \right).
\end{equation}
To obtain the final prompt $\hat{P}$ for each prompt token, we draw $N_s$ independent samples $\{\tilde{P}_k\}_{k=1}^{N_s}$ from the mixture distribution using Eq.~\eqref{eq:sampling}. This improves learning stability by reducing the variance introduced by the stochastic sampling process. However, naively aggregating them via a simple average may fail to reflect the diverse patterns of the query feature, as each sample contributes equally regardless of its relevance to ${q}$. Instead, we aggregate the samples in a way that preserves their diversity while reflecting the characteristics of the query feature, as follows:
\begin{equation} \label{eq:final_prompt}
    \hat{P} = \sum_{k=1}^{N_s} w_k \tilde{P}_k,
\end{equation}
with the aggregation weight $w_k$ defined as follows:
\begin{equation} \label{eq:final_weight}
    w_k  = \frac{\exp \left( \mathrm{sim}(\tilde{P}_k, q)\right)}{\sum_{j=1}^{N_s} \exp \left( \mathrm{sim}(\tilde{P}_j, q) \right)},
\end{equation}
where $\mathrm{sim}(\cdot,\cdot)$ is the cosine similarity. Note that $\hat{P}$ reduces to a simple average when $w_k = 1/N_s$, which we empirically find to be suboptimal~(Sec.~\ref{sec:4.3}). More detailed analyses on the aggregation process can be found in the supplementary material. 

The $M$ prompt tokens obtained from the corresponding pool are then concatenated: 
\begin{equation}
    \hat{\mathbf{P}} = [\hat{P}(1), \dots, \hat{P}(M)] ^{\top} \in \mathbb{R}^{M \times D}. 
\end{equation}
We prepend $\hat{\mathbf{P}}$ to the input by using the Pre-T technique. Note that the sampling process requires only a single forward pass, as we aggregate the $N_s$ samples before applying the Pre-T technique, thus incurring negligible computational and memory costs~(Sec.~\ref{sec:4.3}). We provide an overall sampling algorithm of our approach in the supplementary material.

\vspace{-3mm}
\subsection{Training loss} \label{sec:3.4} 
We train our model with an overall objective, balanced by the regularization parameter of~$\lambda$, as follows:
\begin{equation} \label{eq:total_loss}
    \mathcal{L} = \mathcal{L}_{\mathrm{CE}} + \lambda \mathcal{L}_{\mathrm{DR}},
\end{equation}
where we denote by $\mathcal{L}_{\mathrm{CE}}$ and $\mathcal{L}_{\mathrm{DR}}$ cross-entropy~(CE) and distribution regularization losses, respectively. The distribution regularization term alleviates an abrupt change in the distributions, while optimizing distributions~(or prompts):
\begin{equation} \label{eq:loss}
    \mathcal{L}_{\mathrm{DR}} =  \frac{1}{M} \sum_{m=1}^{M} \frac{1}{N} \sum_{n=1}^{N}\mathcal{L}_{\mathrm{DR}}(m, n), 
\end{equation}
where 
\begin{equation} \label{eq:loss_component}
    \mathcal{L}_{\mathrm{DR}} (m, n) = \text{KL} \left(\mathcal{N} \left( \hat{{\mu}}(m, n),  \hat{{\Sigma}}(m,n)  \right) \, \middle\| \, \mathcal{N} \left( {\mu}(m, n),  {\Sigma}(m,n) \right)\right).
\end{equation}
We denote by $\mathcal{N} ( \hat{{\mu}}(m, n),  \hat{{\Sigma}}(m,n) )$ an $n$-th prompt distribution of an $m$-th pool, similar to~\eqref{eq:dist}, but cached from the previous training step. This regularization term minimizes the Kullback-Leibler~(KL) divergence between prompt distributions over successive training steps to prevent the distributions from changing drastically, which helps to mitigate the forgetting problem.

\vspace{-3mm}
\section{Experiment}

In this section, we describe implementation details~(Sec.~\ref{sec:4.1}) and compare our method with the state of the art on class-incremental image classification~(Sec.~\ref{sec:4.2}). We also provide a detailed analysis of our framework~(Sec.~\ref{sec:4.3}).

\vspace{-3mm}
\subsection{Implementation details} \label{sec:4.1}

\noindent
\textbf{Dataset and evaluation.}
We perform experiments on ImageNet-R~\cite{hendrycks2021many}, CIFAR-$100$~\cite{krizhevsky2009learning}, and CUB-$200$~\cite{wah2011caltech} for class-incremental learning. ImageNet-R consists of $24$K training and $6$K test images for $200$ classes. The classes in ImageNet-R are taken from ImageNet~\cite{deng2009imagenet}, but the images feature different styles, such as sketches, cartoons, and graffiti. Following the standard protocol~\cite{wang2022learning,smith2023coda,wang2022dualprompt,jiao2024vector}, we split ImageNet-R randomly into 5, 10, and 20 disjoint tasks, where each task contains a subset of non-overlapping classes. CIFAR-100~\cite{krizhevsky2009learning} provides 50K training and 10K test images for 100 object categories, and CUB-200~\cite{wah2011caltech} contains 200 bird categories, consisting of 11.8K images. We randomly divide the CIFAR-100 and CUB-200 datasets into 10 tasks, where each task consists of 10 and 20 mutually exclusive classes for CIFAR-100 and CUB200, respectively. Following~\cite{jiao2024vector,wang2023hierarchical,zhang2023slca}, we report final average accuracy~(FAA) and cumulative average accuracy~(CAA) for evaluation. In particular, FAA is the average test accuracy across all tasks after all incremental steps, while CAA is the average FAA, computed at the end of each incremental step. We provide more detailed descriptions of the evaluation metrics in the supplementary material.

\begin{table*}[t]
    \centering
    \caption{
        Quantitative results for 5, 10, and 20 tasks on ImageNet-R~\cite{hendrycks2021many} in terms of FAA and CAA. All numbers are obtained by averaging results over five runs with standard deviations. $\dagger$: The results are reproduced with official codes provided by the authors using a weight pretrained on ImageNet-1K~\cite{russakovsky2015imagenet} for a fair comparison.
    }
    \vspace{-3mm}
    \label{tab:results_imagenet}
    \renewcommand{\arraystretch}{1.05}
    \setlength{\tabcolsep}{0.4pt}
    \scriptsize
    \begin{tabular*}{\textwidth}{@{\extracolsep{\fill}} l c c c c c c}
        \toprule
        \multicolumn{1}{c}{\multirow{2}{*}[-0.6ex]{Method}} 
        & \multicolumn{2}{c}{\textbf{5-task}} 
        & \multicolumn{2}{c}{\textbf{10-task}}  
        & \multicolumn{2}{c}{\textbf{20-task}} \\
        \cmidrule(lr){2-3} \cmidrule(lr){4-5} \cmidrule(lr){6-7} 
        & FAA ($\uparrow$) & CAA ($\uparrow$) 
        & FAA ($\uparrow$) & CAA ($\uparrow$) 
        & FAA ($\uparrow$) & CAA ($\uparrow$) \\
        \midrule \midrule
        Joint-Training                                      &82.06~~~~~~~~~&                
                                                            && 
                                                            &&                 \\
        FT                                                  &18.74 \tiny{$\pm$ 0.44} &48.39 \tiny{$\pm$ 0.58} 
                                                            &10.12 \tiny{$\pm$ 0.51} &35.23 \tiny{$\pm$ 0.92} 
                                                            &~4.75 \tiny{$\pm$ 0.40} &22.80 \tiny{$\pm$ 0.37} \\
        FT++                                                &60.42 \tiny{$\pm$ 0.87} &71.59 \tiny{$\pm$ 0.50}
                                                            &48.93 \tiny{$\pm$ 1.15} &66.79 \tiny{$\pm$ 0.92} 
                                                            &35.98 \tiny{$\pm$ 1.38} &59.68 \tiny{$\pm$ 0.95} \\
        L2P~\cite{wang2022learning}                         &70.83 \tiny{$\pm$ 0.58} &78.34 \tiny{$\pm$ 0.47}
                                                            &69.29 \tiny{$\pm$ 0.73} &78.30 \tiny{$\pm$ 0.69} 
                                                            &65.89 \tiny{$\pm$ 1.30} &77.15 \tiny{$\pm$ 0.65} \\
        L2P++~\cite{wang2022learning}                       &73.93 \tiny{$\pm$ 0.37} &80.14 \tiny{$\pm$ 0.54}
                                                            &71.66 \tiny{$\pm$ 0.64} &79.63 \tiny{$\pm$ 0.90} 
                                                            &68.42 \tiny{$\pm$ 1.20} &78.68 \tiny{$\pm$ 1.03} \\
        DualPrompt~\cite{wang2022dualprompt}                &73.05 \tiny{$\pm$ 0.50} &79.47 \tiny{$\pm$ 0.40}
                                                            &71.32 \tiny{$\pm$ 0.62} &78.94 \tiny{$\pm$ 0.72} 
                                                            &67.87 \tiny{$\pm$ 1.39} &77.42 \tiny{$\pm$ 0.80} \\
        CODA-P~\cite{smith2023coda}                         &76.51 \tiny{$\pm$ 0.38} &82.04 \tiny{$\pm$ 0.54}
                                                            &75.45 \tiny{$\pm$ 0.56} &81.59 \tiny{$\pm$ 0.82} 
                                                            &72.37 \tiny{$\pm$ 1.19} &79.88 \tiny{$\pm$ 1.06} \\
        HiDePrompt~\cite{wang2023hierarchical}   			&76.29 \tiny{$\pm$ 0.10} &78.77 \tiny{$\pm$ 0.11}
                                                            &76.74 \tiny{$\pm$ 0.18} &78.76 \tiny{$\pm$ 0.11}
                                                            &76.46 \tiny{$\pm$ 0.06} &78.76 \tiny{$\pm$ 0.11}\\                            
        EvoPrompt~\cite{kurniawan2024evolving}              &77.16 \tiny{$\pm$ 0.18} &82.22 \tiny{$\pm$ 0.54}
                                                            &76.83 \tiny{$\pm$ 0.08} &82.09 \tiny{$\pm$ 0.68}
                                                            &74.41 \tiny{$\pm$ 0.23} &80.96 \tiny{$\pm$ 1.42}\\
        VQ-Prompt~\cite{jiao2024vector}                     &\ul{79.23} \tiny{$\pm$ 0.29} &{82.96} \tiny{$\pm$ 0.50} 
                                                            &{78.71} \tiny{$\pm$ 0.22} &{83.24} \tiny{$\pm$ 0.68} 
                                                            &\ul{78.10} \tiny{$\pm$ 0.22} &\ul{82.70} \tiny{$\pm$ 1.16} \\ 
        APT$^\dagger$~\cite{chen2025achieving}              & 79.20 \tiny{$\pm$ 0.38} & \ul{83.07} \tiny{$\pm$ 0.45} 
                                                            &\ul{79.05} \tiny{$\pm$ 0.41} & \ul{83.41} \tiny{$\pm$ 0.54} 
                                                            &{75.94} \tiny{$\pm$ 0.04} & 79.46 \tiny{$\pm$ 0.46}\\ 
        \midrule                                                       
        \textbf{{Ours}}                                     &\textbf{{80.53}} \tiny{$\pm$ 0.37}  &\textbf{83.90} \tiny{$\pm$ 0.23} 
                                                            &\textbf{80.23} \tiny{$\pm$ 0.31}  &\textbf{84.21} \tiny{$\pm$ 0.26}
                                                            &\textbf{{79.01}} \tiny{$\pm$ 0.44}  &\textbf{83.58} \tiny{$\pm$ 59} \\   
        \bottomrule
        \vspace{-7mm}
    \end{tabular*}
\end{table*}

\noindent
\textbf{Implementation details.} 
Following the works of~\cite{wang2022learning,wang2022dualprompt,smith2023coda,wang2023hierarchical,kurniawan2024evolving,zhang2023slca,jiao2024vector}, we adopt the ViT-B/16~\cite{dosovitskiy2020image} as our backbone. We use the ViT-B/16 pre-trained on ImageNet-1K~\cite{russakovsky2015imagenet} for ImageNet-R~\cite{hendrycks2021many} and CIFAR-100~\cite{krizhevsky2009learning}, and the ViT-B/16 pre-trained on ImageNet-21K~\cite{ridnik2021imagenet} for CUB-200~\cite{wah2011caltech}. For numerical stability, we parameterize each prompt distribution using its log standard deviation. To train our model, we use the AdamW optimizer~\cite{loshchilov2017decoupled} with $\beta_1$ of 0.9 and $\beta_2$ of 0.999, respectively. The initial learning rate is set to $2.5 \times 10^{-3}$ and scheduled using cosine decay~\cite{loshchilov2016sgdr}. We optimize the prompts and classifiers for 20 epochs across all benchmarks following~\cite{jiao2024vector}. We set the batch size as 128 for CIFAR-100 and CUB-200, and 64 for ImageNet-R. We set the number of prompts sampled per input $N_s$ and the balancing parameter $\lambda$ to $30$ and $1e-6$, respectively, by performing a grid search on separate cross-validation splits on each dataset. Following VQ-Prompt~\cite{jiao2024vector}, we set the number of prompt tokens $M$ and the number distribution $N$ to $8$ and $10$, respectively. We provide a detailed description of hyperparameter settings in the supplementary material. For all experiments, we implement our method using PyTorch and use NVIDIA A5000 GPUs.

\vspace{-4mm}
\subsection{Results} \label{sec:4.2} 

\noindent
\textbf{ImageNet-R.}
We compare in Table~\ref{tab:results_imagenet} our method with other prompt-based continual learning methods for 5, 10, and 20 tasks on ImageNet-R~\cite{hendrycks2021many}. We report average scores over five runs with standard deviations. We reproduce the results for APT~\cite{chen2025achieving} using the official code with the same pre-trained weight for a fair comparison, and the numbers for other methods are taken from VQ-Prompt~\cite{jiao2024vector}. From the table, we can see that our approach outperforms state-of-the-art methods using prompts~\cite{wang2022learning,wang2022dualprompt,smith2023coda,jiao2024vector,wang2023hierarchical} across all tasks by significant margins in terms of FAA and CAA. This indicates that diverse prompts sampled from our method better capture diverse patterns of queries in the continual learning scenarios, which is crucial for prompt-based class-incremental learning. The results also confirm that our distribution regularization loss, preventing the distributions from changing abruptly, alleviates the forgetting problem effectively. 

\begin{table}[t]
    \centering
    \vspace{3mm}
    \caption{Quantitative results for 10 tasks on CIFAR-100~\cite{krizhevsky2009learning} and CUB-200~\cite{wah2011caltech} in terms of FAA and CAA. All numbers are obtained by averaging results over five runs with standard deviations.}
    \vspace{-2mm}
    \label{tab:results_cifar_cub}
    \renewcommand{\arraystretch}{1.05}
    \setlength{\tabcolsep}{6pt}
    \scriptsize
    \begin{tabular}{l cc cc}
        \toprule
        \multirow{2}{*}[-0.6ex]{Method}
        & \multicolumn{2}{c}{\textbf{CIFAR-100 (10-task)}}
        & \multicolumn{2}{c}{\textbf{CUB-200 (10-task)}} \\
        \cmidrule(lr){2-3} \cmidrule(lr){4-5}
        & FAA ($\uparrow$) & CAA ($\uparrow$)
        & FAA ($\uparrow$) & CAA ($\uparrow$) \\
        \midrule \midrule

        Joint-Training
        & 91.38\tiny{~~~~~~~~~~} & -
        & 88.41\tiny{~~~~~~~~~~} & - \\

        FT
        & 29.21 \tiny{$\pm$ 0.18} & 37.37 \tiny{$\pm$ 0.89}
        & 11.04 \tiny{$\pm$ 0.78} & 31.96 \tiny{$\pm$ 0.74} \\

        FT++
        & 49.91 \tiny{$\pm$ 0.42} & 74.76 \tiny{$\pm$ 0.93}
        & 37.81 \tiny{$\pm$ 2.86} & 63.55 \tiny{$\pm$ 1.62} \\

        L2P++~\cite{wang2022learning}
        & 82.50 \tiny{$\pm$ 1.10} & 88.96 \tiny{$\pm$ 0.82}
        & - & - \\

        Deep L2P++~\cite{wang2022learning}
        & 84.30 \tiny{$\pm$ 1.03} & 90.50 \tiny{$\pm$ 0.69}
        & - & - \\

        DualPrompt~\cite{wang2022dualprompt}
        & 79.81 \tiny{$\pm$ 1.19} & 88.48 \tiny{$\pm$ 1.32}
        & 65.01 \tiny{$\pm$ 1.08} & 77.56 \tiny{$\pm$ 0.84} \\

        CODA-P~\cite{smith2023coda}
        & 81.03 \tiny{$\pm$ 0.78} & 84.26 \tiny{$\pm$ 0.84}
        & 73.44 \tiny{$\pm$ 0.62} & 81.55 \tiny{$\pm$ 0.70} \\

        VQ-Prompt~\cite{jiao2024vector}
        & {88.73} \tiny{$\pm$ 0.27} & \ul{92.84} \tiny{$\pm$ 0.73}
        & \ul{86.72} \tiny{$\pm$ 0.94} & \ul{90.33} \tiny{$\pm$ 1.03} \\

        APT~\cite{chen2025achieving}
        & \ul{88.85} \tiny{$\pm$ 0.63} & \ul{92.84} \tiny{$\pm$ 0.59}
        & {78.50} \tiny{$\pm$ 0.94} & - \\

        \midrule
        \textbf{Ours}
        & \textbf{89.38} \tiny{$\pm$ 0.22} & \textbf{93.34} \tiny{$\pm$ 0.51}
        & \textbf{87.52} \tiny{$\pm$ 0.41} & \textbf{90.93} \tiny{$\pm$ 0.70} \\

        \bottomrule
    \end{tabular}
    \vspace{-5mm}
\end{table}

\vspace{1.5mm}
\noindent
\textbf{CIFAR100 and CUB200.}
We also provide in Table~\ref{tab:results_cifar_cub} quantitative results on CIFAR-100~\cite{krizhevsky2009learning} and CUB-200~\cite{wah2011caltech}. We provide average values of FAA and CAA over five runs with standard deviations. For a fair comparison, we reproduce the results of DualPrompt~\cite{wang2022dualprompt}, CODA-P~\cite{smith2023coda}, and APT~\cite{chen2025achieving} using the official codes with the same pre-trained weights described in Sec.~\ref{sec:4.1}, while the numbers for the other methods are taken from VQ-Prompt~\cite{jiao2024vector}. From Table~\ref{tab:results_cifar_cub}, we can see that our method surpasses all previous methods by clear margins in terms of FAA and CAA on both datasets, showing similar trends as in Table~\ref{tab:results_imagenet}. This demonstrates that our framework can handle highly diverse patterns of input images, even for CUB-200~\cite{wah2011caltech}, a fine-grained scenario, where object classes are similar to each other.

\begin{table*}[t]
\centering
\vspace{3mm}
\caption{Quantitative Results for 10 tasks on ImageNet-R~\cite{hendrycks2021many} in terms of FAA and CAA. We adopt ViT-B/16 pre-trained with two self-supervised methods~\cite{zhou2022ibot,caron2021emerging}. Following the works of~\cite{wang2023hierarchical,jiao2024vector}, we provide FAA and CAA of our approach, obtained by averaging results over three runs with standard deviations. Other results are taken from VQ-Prompt~\cite{jiao2024vector}.}
\scriptsize
\label{tab:selfsup}
\renewcommand{\arraystretch}{1.1}
\setlength{\tabcolsep}{5pt}
\begin{tabular}{l c c c c }
\toprule 
\multicolumn{1}{c}{\multirow{2}{*}[-0.6ex]{Method}}  &\multicolumn{2}{c}{\textbf{iBOT-1K}~\cite{zhou2022ibot}} &\multicolumn{2}{c}{\textbf{DINO-1K}~\cite{caron2021emerging}} \\
\cmidrule(lr){2-3} \cmidrule(lr){4-5}
&{FAA ($\uparrow$)} &{CAA ($\uparrow$)} &{FAA ($\uparrow$)} &{CAA ($\uparrow$)}    \\ 
\midrule \midrule

DualPrompt~\cite{wang2022dualprompt}    & 61.51 \tiny{$\pm$ 1.05}      & 67.11 \tiny{$\pm$ 0.08}
                                        & 58.57 \tiny{$\pm$ 0.45}      & 64.89 \tiny{$\pm$ 0.15} \\
CODA-P~\cite{smith2023coda}             & 66.56 \tiny{$\pm$ 0.68}      & 73.14 \tiny{$\pm$ 0.57} 
                                        & 63.15 \tiny{$\pm$ 0.39}      & 69.73 \tiny{$\pm$ 0.25} \\ 
HiDe-Prompt~\cite{wang2023hierarchical} & 71.33 \tiny{$\pm$ 0.21}      & 73.62 \tiny{$\pm$ 0.13} 
                                        & 68.11 \tiny{$\pm$ 0.18}      & 71.70 \tiny{$\pm$ 0.01} \\ 
VQ-Prompt~\cite{jiao2024vector}         & \ul{71.68} \tiny{$\pm$ 0.72}      & \ul{76.66} \tiny{$\pm$ 0.40} 
                                        & \ul{68.42} \tiny{$\pm$ 0.28}      & \ul{74.43} \tiny{$\pm$ 0.58} \\ 
\midrule 
\textbf{Ours}                           &\textbf{72.71} \tiny{$\pm$ 0.32}   & \textbf{ 78.23} \tiny{$\pm$ 0.33} 
                                        &\textbf{69.41} \tiny{$\pm$ 0.39}   & \textbf{ 75.50} \tiny{$\pm$ 0.37} \\

\bottomrule
\end{tabular}
\vspace{-6mm}
\end{table*}

\vspace{-3mm}
\subsection{Discussion} \label{sec:4.3}

\vspace{1mm}
\noindent
\textbf{Comparison of pre-training regimes.}
We show in Table~\ref{tab:selfsup} quantitative results for 10 tasks on ImageNet-R~\cite{hendrycks2021many}, adopting ViT-B/16 pre-trained on iBOT-1K~\cite{zhou2022ibot} and DINO-1K~\cite{caron2021emerging}. From Table~\ref{tab:selfsup}, we can see that our method outperforms previous methods by clear margins in terms of FAA and CAA for both pre-trained weights, suggesting that our approach works favorably for different pre-training paradigms and it can generalize well across diverse initialization settings. 

\begin{figure*}[t]
    \vspace{3mm}
    \centering
    \begin{subfigure}{0.193\textwidth}
        \centering
        \includegraphics[width=\textwidth]{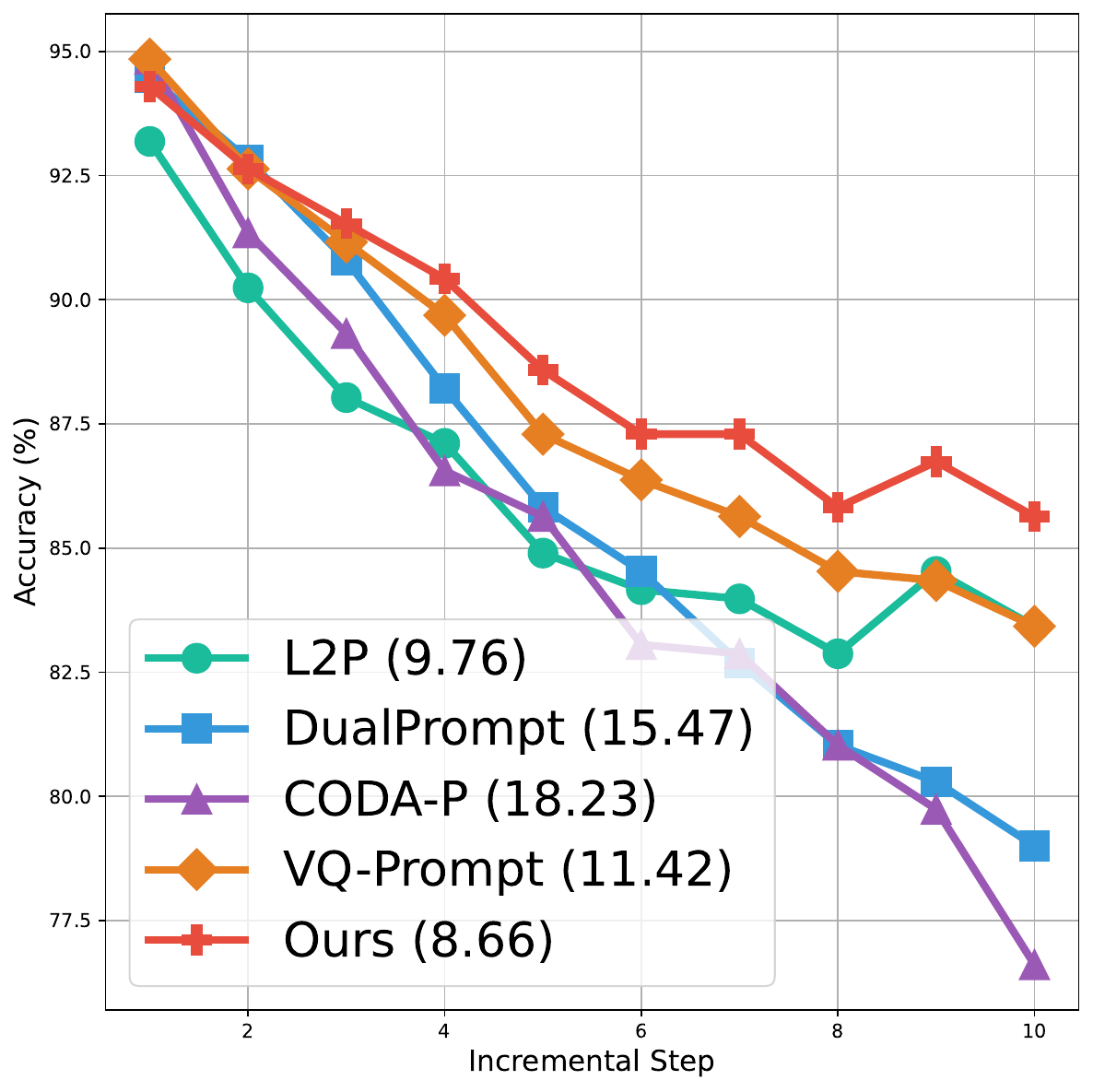}
        \caption*{~~~~~Task 1.}
    \end{subfigure}
    \begin{subfigure}{0.193\textwidth}
        \centering
        \includegraphics[width=\textwidth]{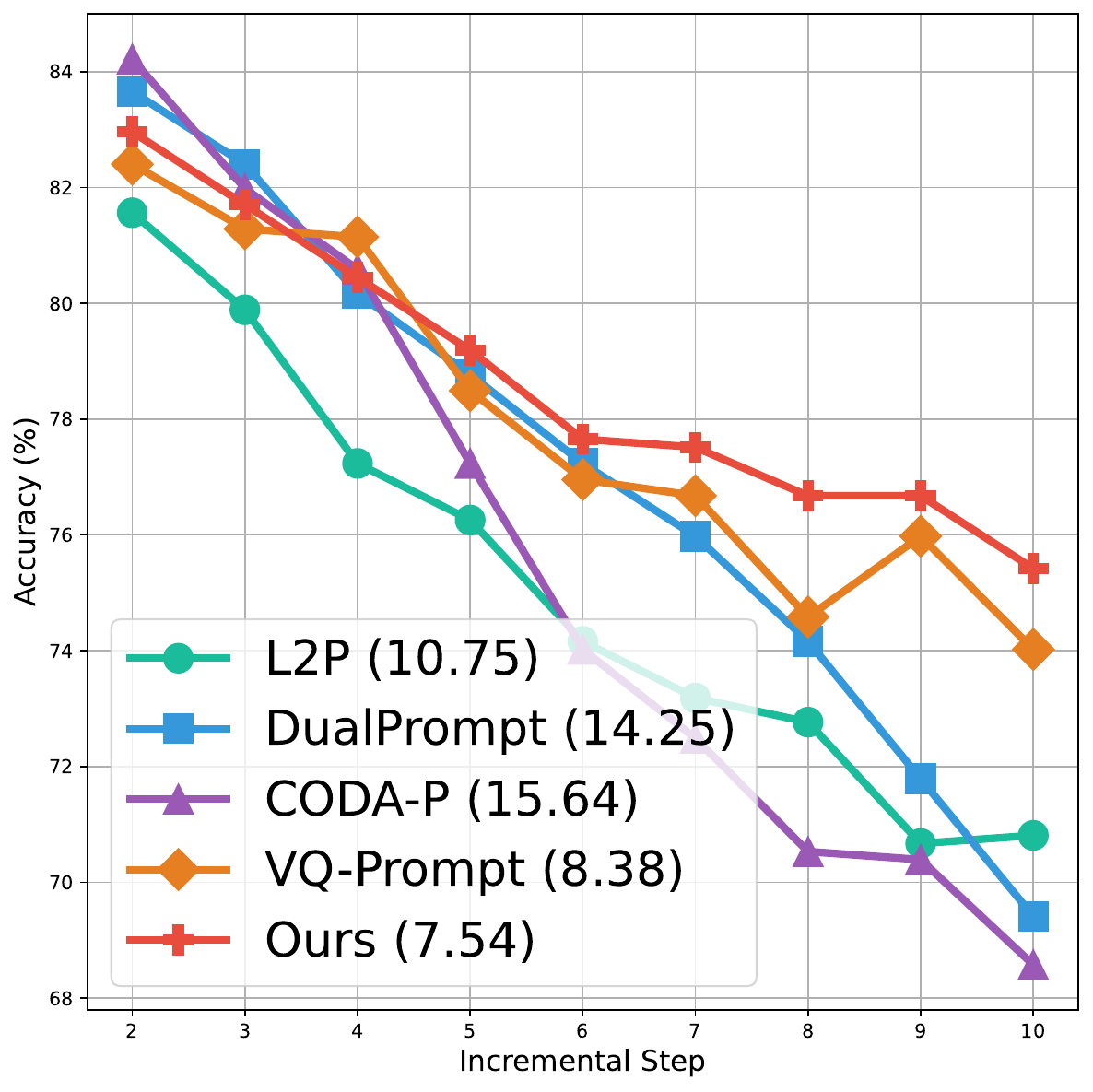}
        \caption*{~~~~~Task 2.}
    \end{subfigure}
    \begin{subfigure}{0.193\textwidth}
        \centering
        \includegraphics[width=\textwidth]{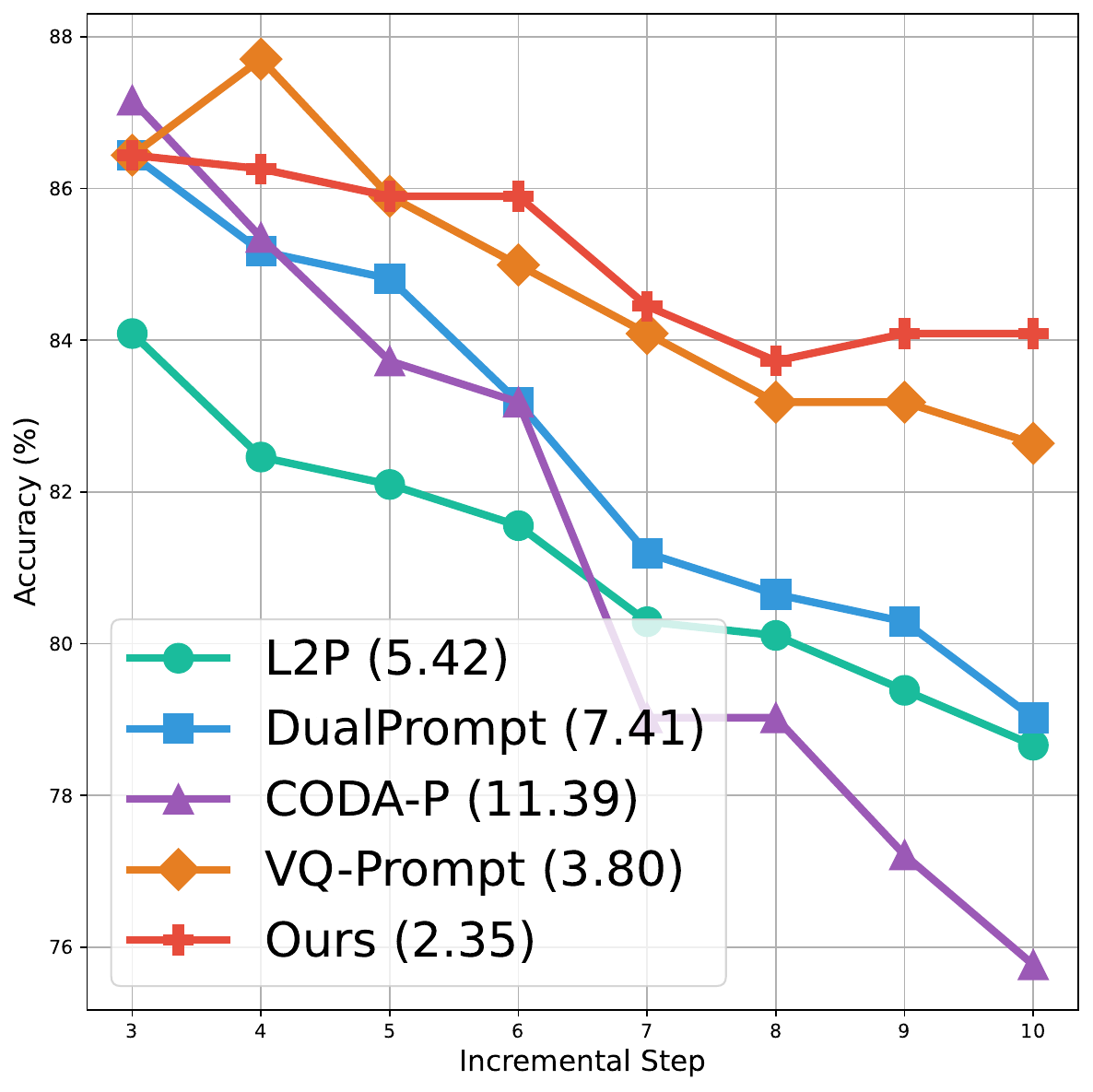}
        \caption*{~~~~~Task 3.}
    \end{subfigure}
    \begin{subfigure}{0.193\textwidth}
        \centering
        \includegraphics[width=\textwidth]{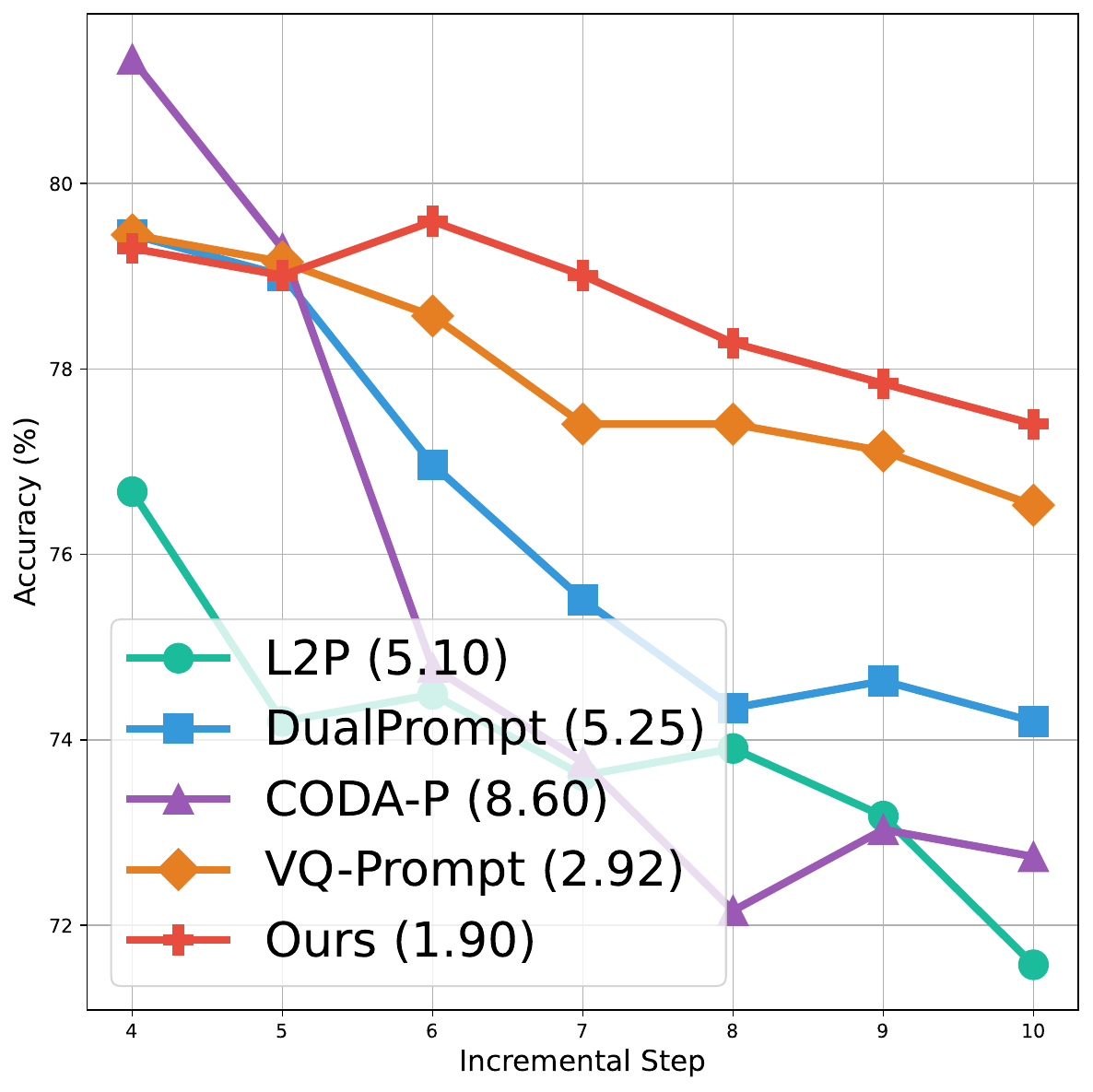}
        \caption*{~~~~~Task 4.}
    \end{subfigure}
    \begin{subfigure}{0.193\textwidth}
        \centering
        \includegraphics[width=\textwidth]{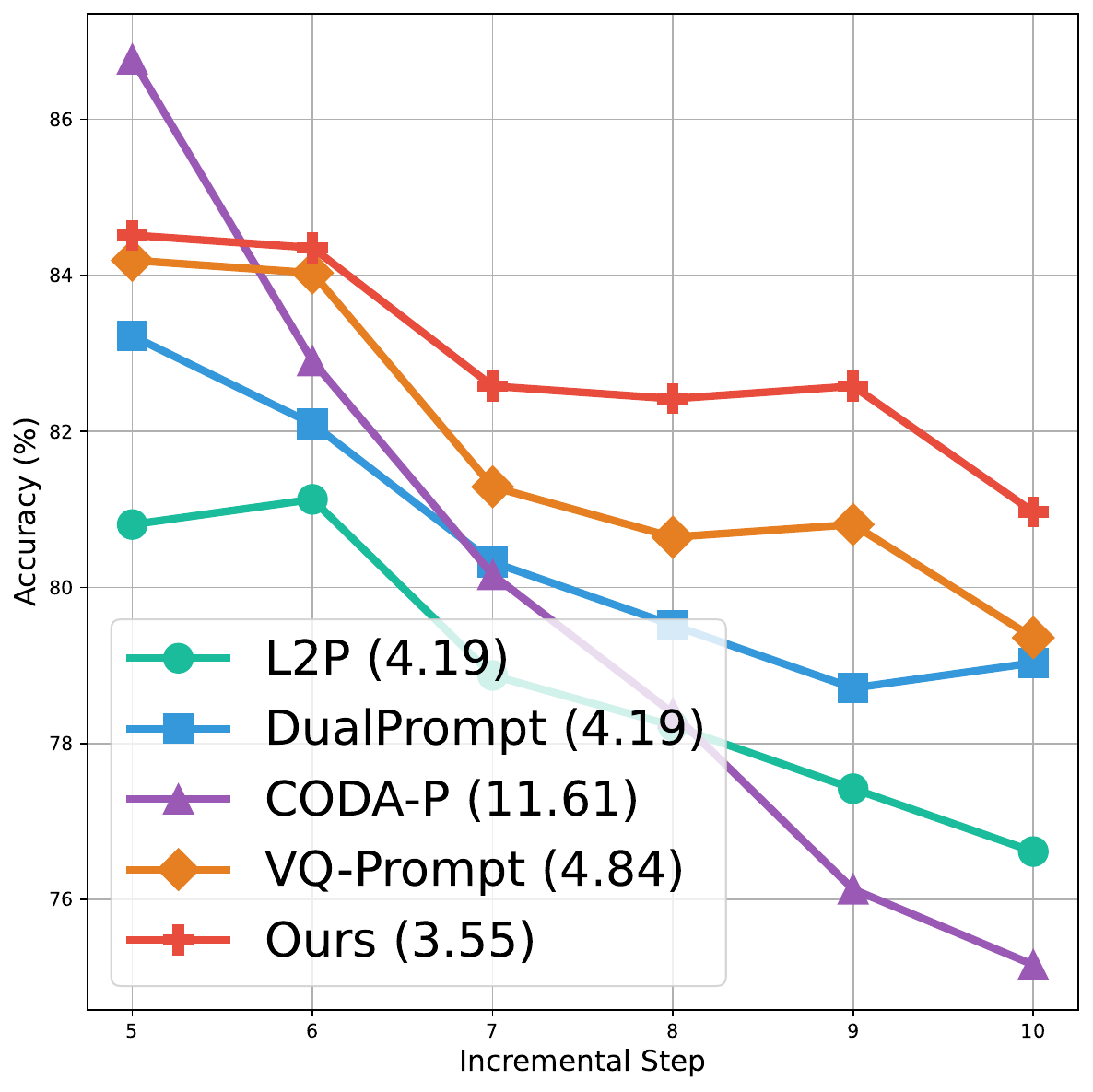}
        \caption*{~~~~~Task 5.}
    \end{subfigure}

    \vspace{1mm}
    \begin{subfigure}{0.193\textwidth}
        \centering
        \includegraphics[width=\textwidth]{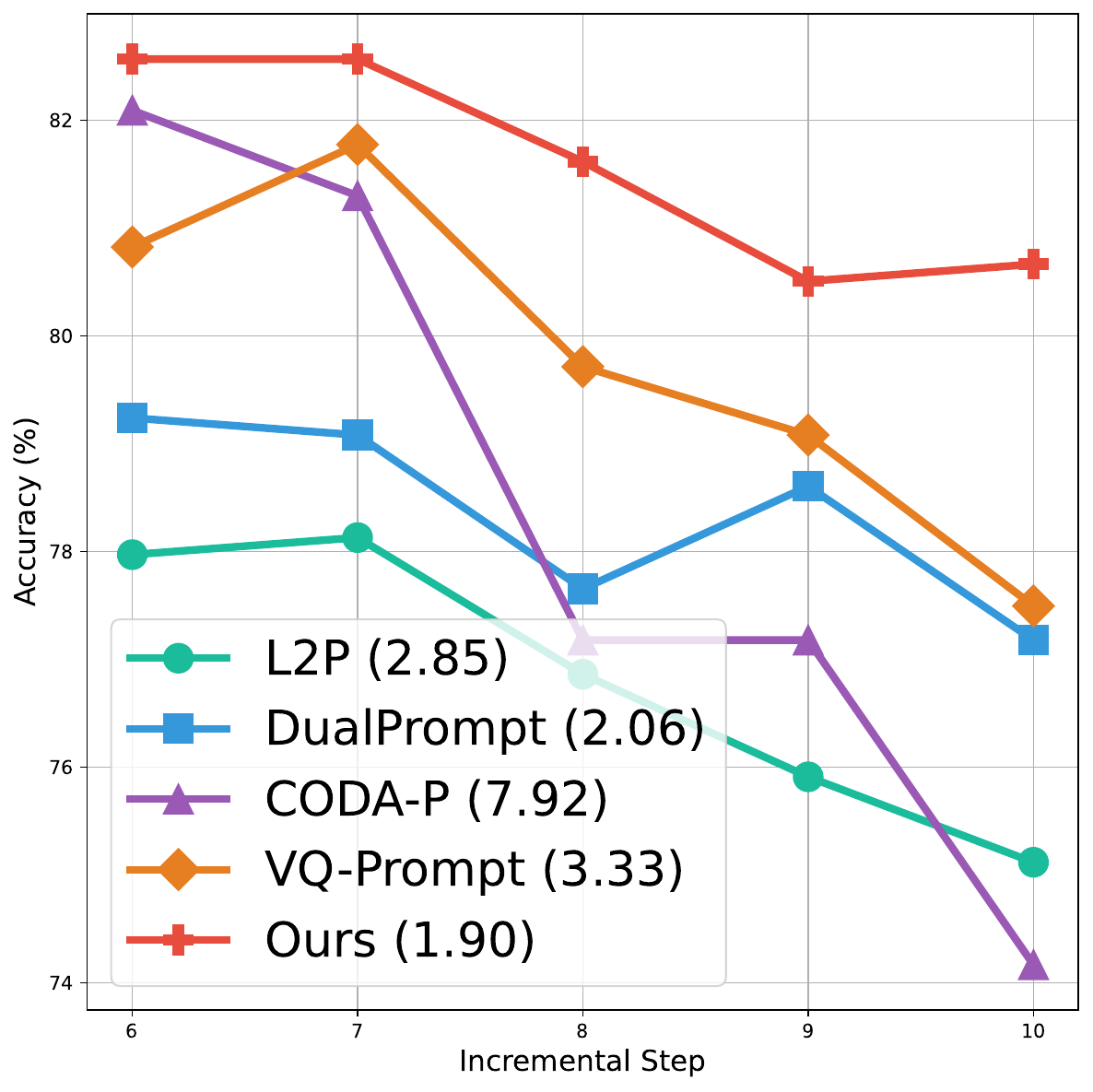}
        \caption*{~~~~~Task 6.}
    \end{subfigure}
    \begin{subfigure}{0.193\textwidth}
        \centering
        \includegraphics[width=\textwidth]{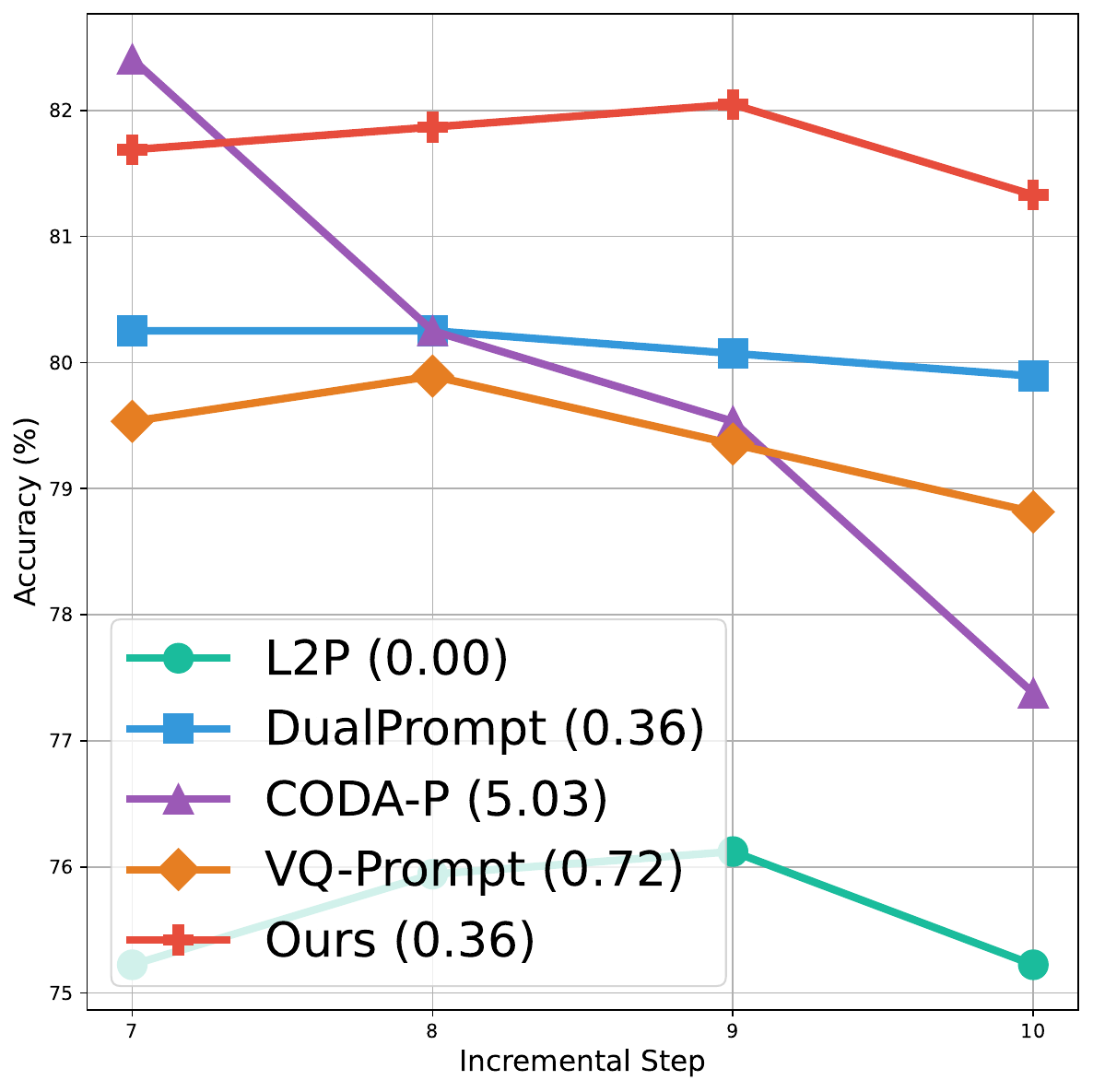}
        \caption*{~~~~~Task 7.}
    \end{subfigure}
    \begin{subfigure}{0.193\textwidth}
        \centering
        \includegraphics[width=\textwidth]{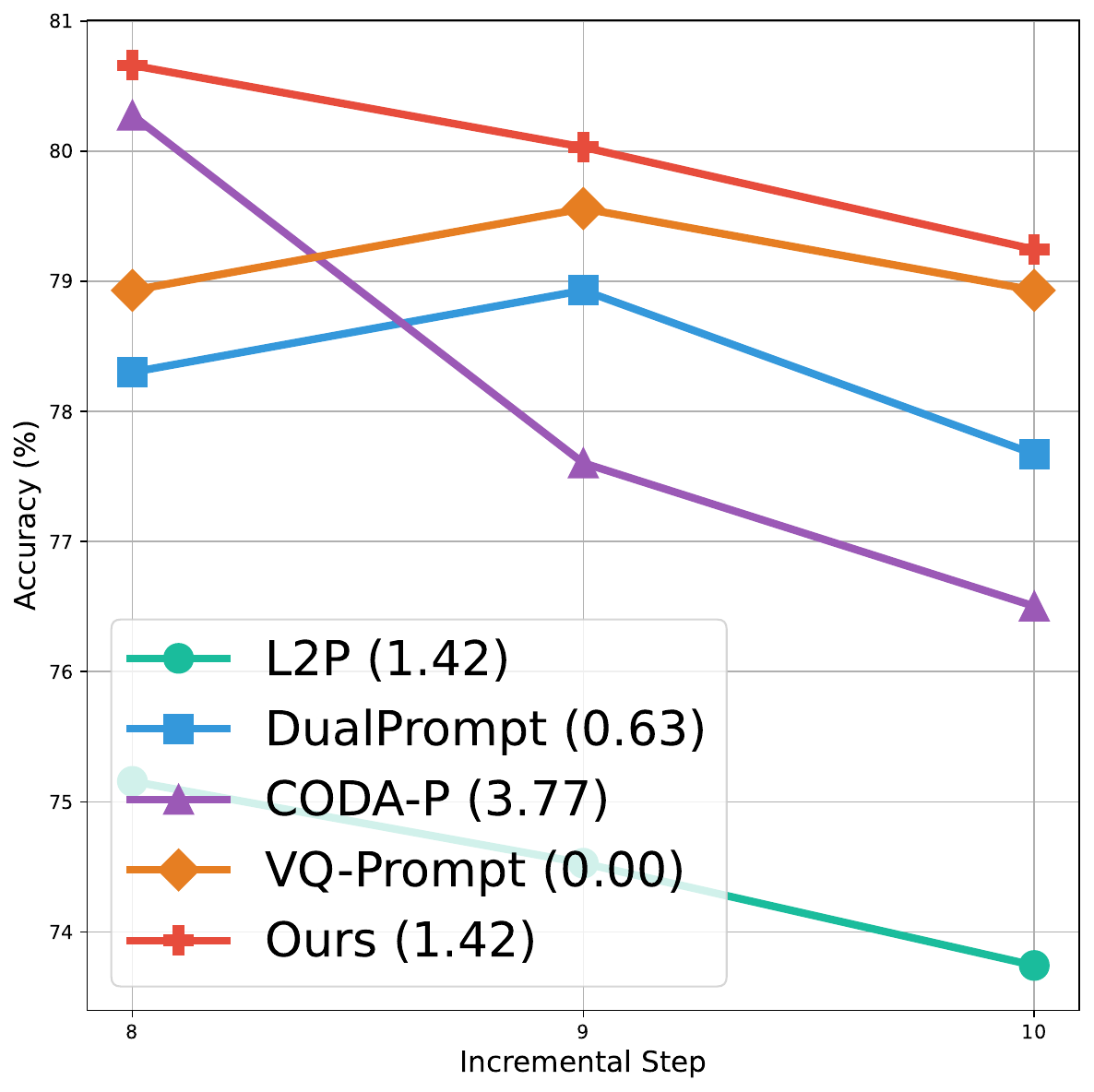}
        \caption*{~~~~~Task 8.}
    \end{subfigure}
    \begin{subfigure}{0.193\textwidth}
        \centering
        \includegraphics[width=\textwidth]{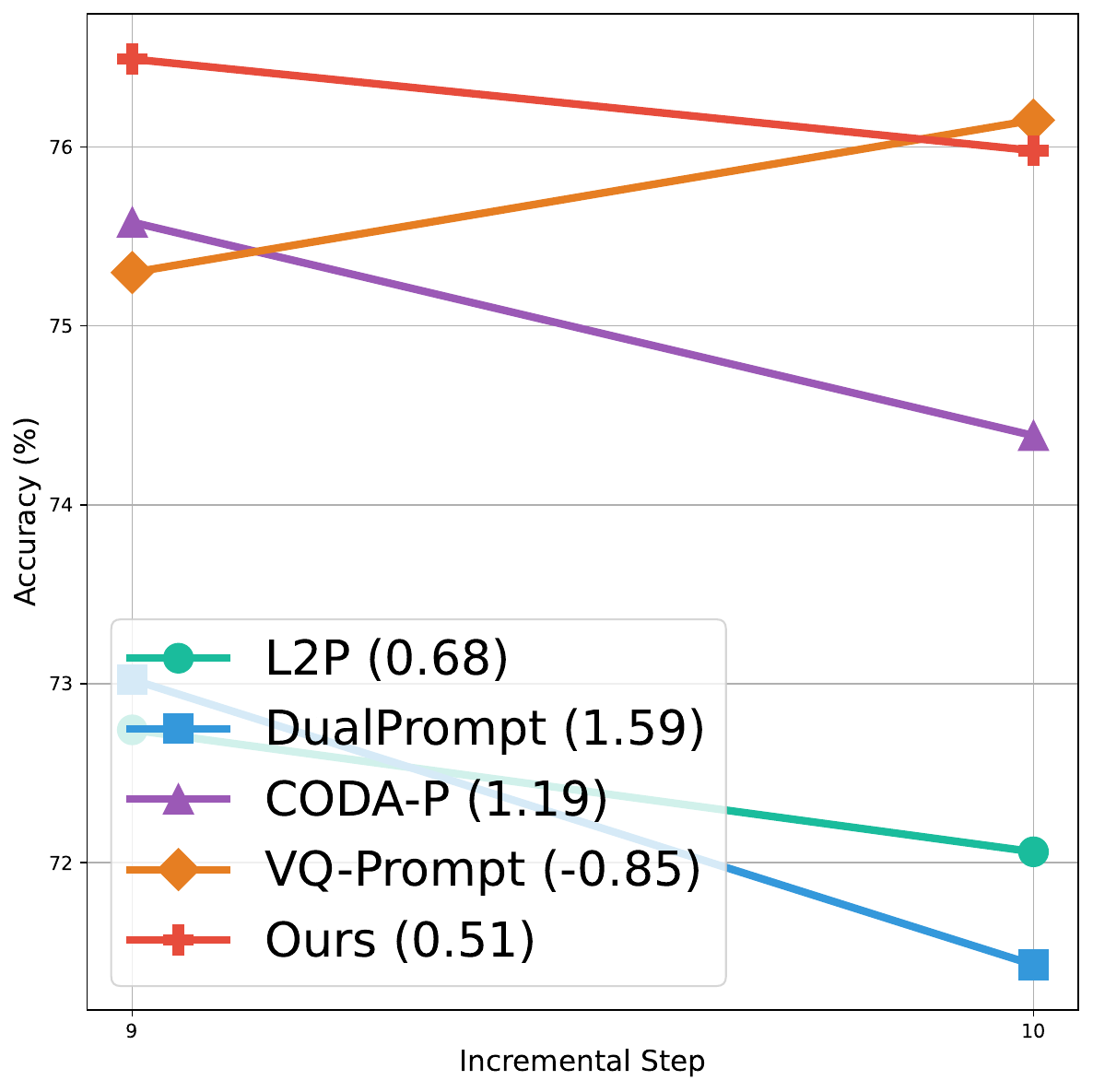}
        \caption*{~~~~~Task 9.}
    \end{subfigure}
    \begin{subfigure}{0.193\textwidth}
        \centering
        \includegraphics[width=\textwidth]{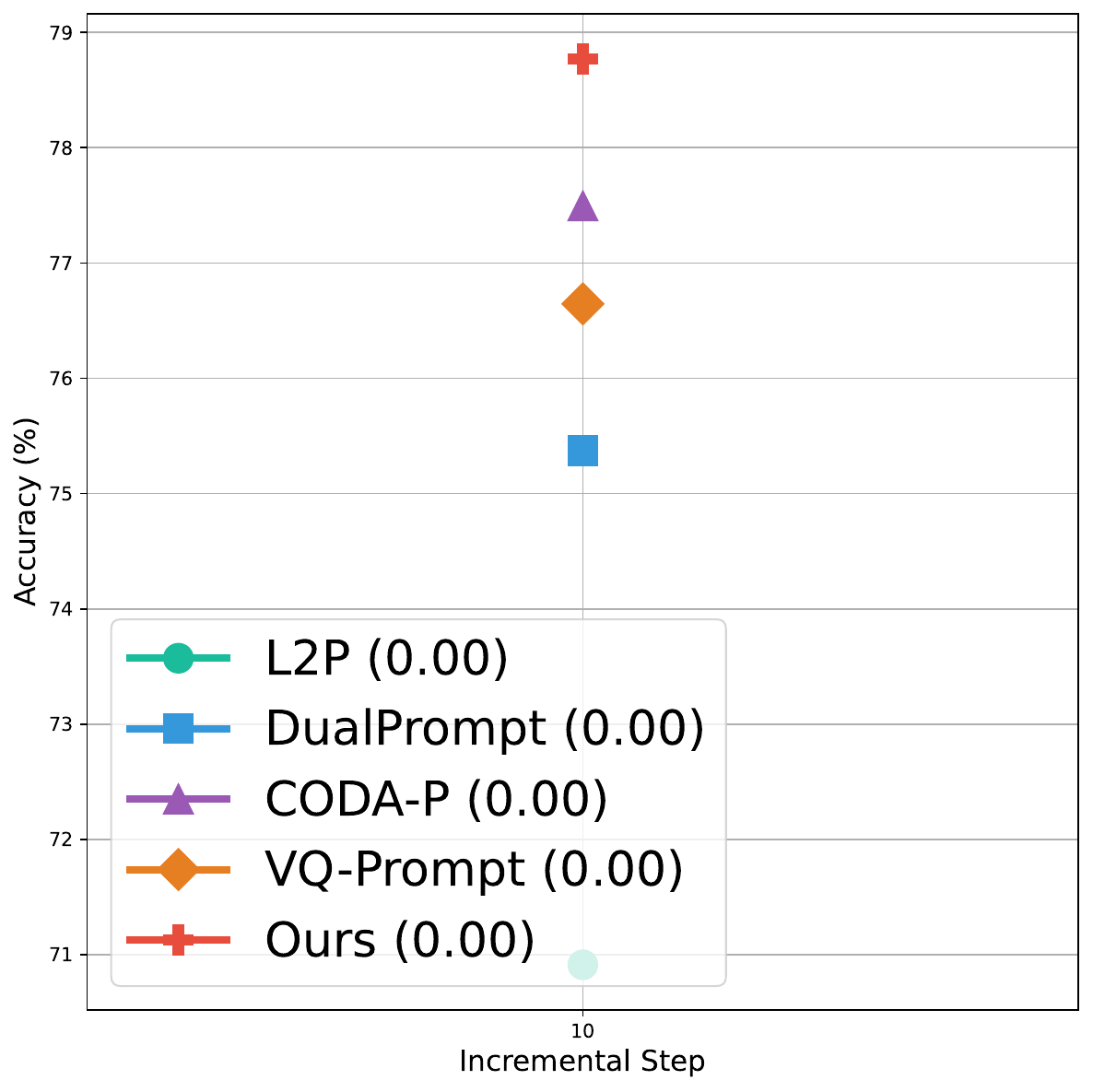}
        \caption*{~~~~~Task 10.}
    \end{subfigure}
    \vspace{-4mm}
    \caption{Per-task accuracy over incremental steps for the 10-task split on ImageNet-R~\cite{hendrycks2021many}. It shows the accuracy of each task, from the initial learning step to the final incremental step. Numbers in parentheses in the legend indicate the accuracy drop for each method, that is, the difference between the accuracies at the initial and final steps, where lower values indicate that the model better preserves previously learned knowledge. Best viewed by zooming in with color. We provide more results in the supplementary material.}
    \label{fig:step}
    \vspace{-7mm}
\end{figure*}

\noindent
\textbf{Per-task accuracy over incremental steps.}
We show in Fig.~\ref{fig:step} per-task accuracy for the 10 tasks split on ImageNet-R~\cite{hendrycks2021many}. From the figure, we can see that our method achieves higher accuracy across most tasks over the incremental steps. This is because diverse prompts enable the model to capture diverse image patterns in a sequence of data effectively. Additionally, the accuracy drop after incremental steps in our method is generally smaller than that of other methods, indicating that the distribution regularization loss addresses the forgetting problem effectively. These two observations demonstrate that our method learns new tasks effectively, while retaining previously acquired knowledge.

\vspace{1.mm}
\begin{wraptable}{r}{0.6\textwidth} 
\centering
\vspace{-12pt}
\caption{Ablation analyses for 10 tasks on ImageNet-R~\cite{hendrycks2021many}. $\mathcal{N}$: Using a prompt distribution in Eq.~\eqref{eq:dist}. $\mathcal{N}_{\mathrm{GM}}$: Using a mixture of prompt distributions in Eq.~\eqref{eq:mixture} and combining prompts via Eq.~\eqref{eq:final_prompt}.}
\label{tab:ablation}
\scriptsize
\renewcommand{\arraystretch}{1.1}
\setlength{\tabcolsep}{4pt}
\begin{tabular}{ccc cc}
\toprule
$\mathcal{N}$ & $\mathcal{N}_{\mathrm{GM}}$ & $\mathcal{L}_{\mathrm{DR}}$ & FAA ($\uparrow$) & CAA ($\uparrow$) \\
\midrule \midrule
           &                &            & 77.49 \tiny{$\pm$ 0.27}  & 80.88 \tiny{$\pm$ 0.43}  \\
\checkmark &                &            & 78.73 \tiny{$\pm$ 0.51}  & 82.46 \tiny{$\pm$ 0.47}  \\
\checkmark & \checkmark     &            & \ul{79.98} \tiny{$\pm$ 0.49}  & 83.63 \tiny{$\pm$ 0.39}  \\
\checkmark &                & \checkmark & 79.79 \tiny{$\pm$ 0.63}  & \ul{83.80} \tiny{$\pm$ 0.42}  \\
\checkmark &  \checkmark    & \checkmark & {\textbf{80.23}} \tiny{$\pm$ 0.31}  & \textbf{84.21} \tiny{$\pm$ 0.26}  \\
\bottomrule
\end{tabular}
\vspace{2pt}
\end{wraptable}
\noindent
\textbf{Ablation study.}
We provide in Table~\ref{tab:ablation} an ablation study on each component of our method. As a baseline in the first row, we use the mean vector $\mu$ alone with unit variance, \emph{e.g.}, $\Sigma=\mathbf{I}$, where $\mathbf{I}$ is a $D \times D$ identity matrix. This baseline corresponds to a deterministic variant of our method, where the sampling process is removed while the other components remain unchanged. For the variant in the second row, we aggregate prompts, sampled from each distribution independently, using the relevance scores in Eq.~\eqref{eq:score} without forming a mixture. From the table, we have three findings as follows:~(1) The first and second rows show that modeling prompts as probabilistic distributions provides better results. A plausible reason is that the stochasticity of the sampling process allows us to facilitate diverse prompts, which enables our model to capture various patterns in input images.~(2) From the second and third rows, we can observe that our framework, that leverages a mixture of prompt distributions, performs better. This is because aggregating prompts, sampled from each distribution independently, may neglect the relations between distributions and input tokens.~(3) From the last four rows, we can see that our distribution regularization loss mitigates the forgetting problem effectively by avoiding an abrupt change in distributions.

\begin{wraptable}{r}{0.5\textwidth} 
    \centering
    \vspace{-11pt}
    \caption{Comparison of FAA w.r.t the number of selected distributions $K$ for the 5, 10, and 20 tasks on ImageNet-R~\cite{hendrycks2021many}.}
    \label{tab:topk}
    \scriptsize
    \renewcommand{\arraystretch}{1.1}
    \setlength{\tabcolsep}{6pt}
    \begin{tabular}{c ccc}
        \toprule
        $K$ & 5-task & 10-task & 20-task \\
        \midrule \midrule
        7  & 80.86 & 80.21 & 79.21 \\
        8  & 80.79 & 80.41 & 79.40 \\
        9  & 80.70 & 80.43 & 79.34 \\
        10 & 80.68 & 80.35 & 79.46 \\
        \bottomrule
    \end{tabular}
    \vspace{2pt}
\end{wraptable}
\noindent
\textbf{Effect of irrelevant distributions.} The relevance scores in Eq.~\eqref{eq:score} assign non-zero weights to all prompt distributions, which could, in principle, allow irrelevant distributions to affect the mixture. However, the softmax in Eq.~\eqref{eq:score} can suppress such distributions. To verify this, we show in Table~\ref{tab:topk} the results of constructing the mixture using only the $K$ most relevant distributions, while discarding the others, on ImageNet-R~\cite{hendrycks2021many}. Note that $K=N$ corresponds to our method in Eq.~\eqref{eq:mixture}. The results show that discarding low-relevance distributions provides only marginal gains over our soft mixture, suggesting that the softmax suppresses them effectively. Meanwhile, the optimal $K$ varies across task splits even within the same dataset. This introduces an additional scenario-dependent hyperparameter, which is particularly impractical in continual learning, where task distributions change over time and future tasks are unknown. We thus retain the soft mixture to avoid scenario-specific tuning, while achieving comparable performance.

\vspace{1.5mm}
\begin{wraptable}{r}{0.55\textwidth} 
    \centering
    \vspace{-12pt}
    \caption{Comparison of FAA and CAA w.r.t the number of samples, $N_s$, for the 10 tasks on ImageNet-R~\cite{hendrycks2021many}. We average the results over five runs with standard deviations. $-$: Using $\mu_{\mathrm{GM}}$ without sampling.}
    \label{tab:ns_effect_training}
    \vspace{1pt}
    \renewcommand{\arraystretch}{1.1}
    \setlength{\tabcolsep}{6pt}
    \scriptsize
    \begin{tabular}{c cc}
        \toprule 
        $N_s$ & FAA ($\uparrow$) & CAA ($\uparrow$) \\
        \midrule \midrule
        1     & Fail to converge &   \\
        5     & 33.55 \tiny{$\pm$ 13.61} & 36.57 \tiny{$\pm$ 11.24} \\
        10    & 61.04 \tiny{$\pm$ 7.19} & 62.16 \tiny{$\pm$ 8.05} \\
        20    & \ul{80.00} \tiny{$\pm$ 0.38} & {83.97} \tiny{$\pm$ 0.41} \\
        30    & \textbf{80.23} \tiny{$\pm$ 0.31} & \textbf{84.21} \tiny{$\pm$ 0.26} \\
        40    & 79.96 \tiny{$\pm$ 0.28} & \ul{84.01} \tiny{$\pm$ 0.22} \\
        $-$ & 79.18 \tiny{$\pm$ 0.18} & {82.54} \tiny{$\pm$ 0.20} \\
        \bottomrule
    \end{tabular}
    \vspace{3pt}
\end{wraptable}
\noindent
\textbf{Effect of $N_s$.}
To train our model, we sample multiple prompts from the learned mixture distribution and combine them using Eq.~\eqref{eq:final_prompt}. We show in Table~\ref{tab:ns_effect_training} the effect of the number of sampled prompts, denoted by $N_s$, in terms of FAA and CAA. 
From the table, we have three findings as follows:~(1)~From the first three rows, we can see that training becomes unstable when $N_s$ is small,~\emph{e.g.}, $N_s = 1$, $5$, and $10$. This is because the randomness from insufficient sampling hinders our model from learning task information. In particular, the model fails to converge if we use a single prompt.~(2)~The fourth to sixth rows show that using a moderate number of prompts, such as $N_s=20$, $30$, or $40$, enables our model to achieve stable training and better performance than the previous methods~\cite{wang2022dualprompt,wang2022learning,smith2023coda,jiao2024vector,wang2023hierarchical,kurniawan2024evolving}. This indicates that the prompts from our method benefit from stochasticity in the sampling process, which helps to learn rich information of data distributions in the continaul learning process (3) We can see from the last row that using $\mu_{\mathrm{GM}}$ yields reasonable, but worse results, compared to the cases of $N_s$ of 20, 30, and 40, as the model is still exploring diverse prompts to represent various input images during training. This suggests that exploiting $\mu_{\mathrm{GM}}$ at training time is sub-optimal, confirming once more the importance of stochastic sampling during training. We provide a more detailed analysis on the effect of $N_s$ over a wider range of values in the supplementary material.

\begin{table}[t]
\centering
\vspace{6mm}
\caption{Comparisons of FAA for the 10 tasks on ImageNet-R~\cite{hendrycks2021many}, the memory incurred by each prompt during a single forward pass, the training time per epoch, and the inference speed.}
\vspace{-1mm}

\label{tab:runtime_imr_r2}
\scriptsize
\renewcommand{\arraystretch}{1.2}
\setlength{\tabcolsep}{2pt}
\begin{tabular}{l c c c c}
\hline
\multicolumn{1}{c}{\multirow{2}{*}{Method}}  &
\multicolumn{1}{c}{FAA} &
\multicolumn{1}{c}{Memory} &
\multicolumn{1}{c}{Training} &
\multicolumn{1}{c}{Inference}  \\
& \multicolumn{1}{c}{($\uparrow$)} &
  \multicolumn{1}{c}{(MB)} &
  \multicolumn{1}{c}{time (min)} &
  \multicolumn{1}{c}{speed (FPS)} \\
\hline\hline
L2P~\cite{wang2022learning}             & 69.29 \tiny{$\pm$ 0.73}        & 342.76 & 5.64  & 74.54 \\
DualPrompt~\cite{wang2022dualprompt}    & 71.32 \tiny{$\pm$ 0.62}          & 346.88 & 5.82  & 73.97 \\
CODA-P~\cite{smith2023coda}             & 75.45 \tiny{$\pm$ 0.56}          & 352.59 & 6.09 & 74.38 \\
VQ-Prompt~\cite{jiao2024vector}         & {78.71} \tiny{$\pm$ 0.22}           & 343.87 & 6.30 & {74.15} \\
Ours                                    & 80.23 \tiny{$\pm$ 0.31}          & 347.85 & 6.33  & 74.03 \\
\hline
\end{tabular}
\vspace{-4mm}
\end{table}

\vspace{1.5mm}
\noindent
\begin{wraptable}{r}{0.55\textwidth} 
    \centering
    \vspace{-11pt} 
    \caption{Comparison of FAA and CAA w.r.t. the aggregation schemes.}
    \label{tab:agg_comparison}
    \scriptsize
    \renewcommand{\arraystretch}{1.2}
    \setlength{\tabcolsep}{2pt}
    \begin{tabular}{c cc}
        \toprule
        Eq~\eqref{eq:final_weight} & FAA ($\uparrow$) & CAA ($\uparrow$) \\
        \midrule        \midrule
         & 79.06\tiny{$\pm$0.58} & 83.21\tiny{$\pm$0.51} \\
        \checkmark & \textbf{79.98}\tiny{$\pm$0.49} & \textbf{83.63}\tiny{$\pm$0.39} \\
        \bottomrule
    \vspace{1pt}
    \end{tabular}
\end{wraptable}
\textbf{Aggregation scheme.} 
We aggregate multiple samples using Eq.~\eqref{eq:final_weight} to obtain prompts. To verify the effectiveness of the aggregation strategy, we compare it with a baseline that averages the samples, that is, $w_k=1/N_s$. We show in Table~\ref{tab:agg_comparison} the performance comparison between the two sampling strategies in terms of FAA and CAA without $\mathcal{L}_{\mathrm{DR}}$. From the table, we can observe that the aggregation using Eq.~\eqref{eq:final_weight} outperforms the simple average counterpart by significant margins. This suggests that a simple average, which treats all samples equally, fails to consider the query feature properly due to the inherent stochasticity of the sampling process. In contrast, our aggregation scheme in Eq.~\eqref{eq:final_weight} prioritizes each sample adaptively based on the relation between the sampled prompts and a given query feature, allowing the final prompt to reflect the diverse patterns present in a sequence of data while focusing on the query feature.

\vspace{1.5mm}
\noindent
\textbf{Computational and memory costs.} 
We show in Table~\ref{tab:runtime_imr_r2} the computational and memory costs, including the memory incurred by each prompt during a single forward pass, inference speed, and training time per epoch of ours and previous methods~\cite{wang2022learning,wang2022dualprompt,smith2023coda,jiao2024vector}. From the table, we can see that our method has comparable computational and memory costs to the previous works, while outperforming them by a clear margin. This is because we average sampled prompts before applying the Pre-T technique, enabling a single forward pass per input, rather than $N_s$ forward passes, and adding only negligible computational and memory cost. Furthermore, our framework maintains a memory footprint comparable to previous works~\cite{wang2022learning,wang2022dualprompt,smith2023coda,jiao2024vector}, because we do not use key~\cite{wang2022learning,wang2022dualprompt,smith2023coda,jiao2024vector} or learnable attention matrices~\cite{smith2023coda}, while introducing additional parameters for $\Sigma$.

\subsection{Limitations and Future Work}

Our framework effectively addresses the prompt collapse problem and outperforms existing methods across all benchmarks. Nevertheless, our method is currently evaluated using ViT-based pre-trained backbones on standard class-incremental learning benchmarks. Extending it to more realistic settings, such as open-world continual learning, and to diverse network architectures would be a promising direction for future work.

\section{Conclusion}

We have identified the prompt collapse problem prevalent in existing prompt-based continual learning methods, where the prompts tend to be highly similar to each other and fail to represent the diverse data distributions in the continual learning process. To address this, we have introduced a novel prompt-based framework that models each prompt as a probabilistic distribution and samples diverse prompts from a query-conditioned mixture of the learned distributions, enabling the model to effectively capture the diverse data distributions in the sequence of tasks. We have also presented a distribution regularization loss that prevents the prompt distributions from changing abruptly during the continual learning process, mitigating the forgetting problem. Extensive experimental results demonstrate that our framework better captures the diverse patterns of images than existing prompt-based methods, consistently outperforming them across all benchmarks by clear margins.

\section*{Acknowledgements}
This work was partly supported by IITP grant funded by the Korea government (MSIT) (No.RS-2025-09942968, AI Semiconductor Innovation Lab~(Yonsei University), No.RS-2022-00143524, Development of Fundamental Technology and Integrated Solution for Next-Generation Automatic Artificial Intelligence System, and No.2022-0-00124, RS-2022-II220124, Development of Artificial Intelligence Technology for Self-Improving Competency-Aware Learning Capabilities).

\bibliographystyle{splncs04}
\bibliography{main}

\clearpage


\section*{Supplementary material}

# Learning Probabilistic Prompt for Continual Learning Supplement

Hyekang Park[1], Sanghoon Lee[1], Geon Lee[1], Jongyoun Noh[2], and Bumsub Ham[1⋆]
http://cvlab.yonsei.ac.kr/projects/ProbPrompt

[1]Yonsei University, Seoul, South Korea [2]Samsung Electronics, Suwon, South Korea

In this supplementary material, we describe more details on the experimental settings and hyperparameters (Sec. A). We also provide more results (Sec. B) and discussions (Sec. C), including additional results and observations.

## A Implementation details

### A.1 Metrics

Following the standard protocol in [7,16,17,19], we report final average accuracy (FAA) and cumulative average accuracy (CAA). FAA is defined as follows:

$$\text{FAA} = \frac{1}{T}\sum_{i=1}^{T} \text{Acc}_{i,T}, \tag{i}$$

where we denote by $\text{Acc}_{i,T}$ an accuracy of task $i$ after incremental step $T$. Note that $i \in \{1, \dots T\}$ is an index of task, where $T$ is the number of tasks. FAA is the average test accuracy across all tasks at the end of all incremental steps. That is, FAA shows the capability of models to learn new concepts continually, while preserving previous knowledge. We also use CAA, defined as follows:

$$\text{CAA} = \frac{1}{T}\frac{1}{i}\sum_{i=1}^{T}\sum_{j=1}^{i} \text{Acc}_{j,i}. \tag{ii}$$

To measure CAA, we first compute FAA after each incremental step, and then average the FAA values across an entire task. Similar to FAA, CAA is the ability of models to retain the knowledge they have learned. We also report in this material the forgetting ratio ($F$) to measure how much the models forget previously acquired knowledge after learning a new task. To this end, we compute $F$ as follows:

$$F = \frac{1}{T-1}\sum_{i=1}^{T-1}\left(\max_{j\in\{1,\dots,i\}} \text{Acc}_{j,i} - \text{Acc}_{T,i}\right). \tag{iii}$$

$F$ shows the average performance degradation of previous tasks after learning an entire task. A lower value of $F$ indicates that the model mitigates the forgetting problem and retains prior knowledge more effectively.

⋆ Corresponding author.

**Table A:** Hyperparameter settings.

| Dataset | Task | $N$ | $M$ | $N_s$ | $\lambda$ |
|---|---|---|---|---|---|
| ImageNet-R [4] | **5** | 10 | 8 | 30 | $1e-6$ |
| | **10** | 10 | 8 | 30 | $1e-6$ |
| | **20** | 10 | 8 | 30 | $1e-6$ |
| CUB-200 [15] | **10** | 10 | 8 | 30 | $1e-6$ |
| CIFAR-100 [9] | **10** | 10 | 8 | 30 | $1e-6$ |

**Table B:** Effect of $N$ for 10 tasks on ImageNet-R [4].

| $M$ | FAA (↑) | CAA (↑) |
|---|---|---|
| 20 | 79.90 | 82.78 |
| 16 | 80.13 | 83.15 |
| 12 | 80.37 | 83.63 |
| 8 | 80.35 | 83.88 |
| 4 | 79.23 | 83.11 |
| 1 | 70.75 | 81.42 |

**Table C:** Effect of $N$ for 10 tasks on ImageNet-R [4].

| $N$ | FAA (↑) | CAA (↑) |
|---|---|---|
| 25 | 80.29 | 83.61 |
| 20 | 80.36 | 83.83 |
| 15 | 80.42 | 84.02 |
| 10 | 80.35 | 83.88 |
| 5 | 79.58 | 83.12 |
| 1 | 68.46 | 79.40 |

**Table D:** Effect of $\lambda$ for 10 tasks on ImageNet-R [4] in terms of FAA, CAA, and $F$.

| $\lambda$ | FAA (↑) | CAA (↑) | $F$ (↓) |
|---|---|---|---|
| 5e − 5 | 79.75 | 83.17 | 2.52 |
| 1e − 5 | 79.94 | 83.64 | 2.58 |
| 5e − 6 | 80.33 | 83.96 | 2.65 |
| 1e − 6 | 80.35 | 83.88 | 2.91 |
| 5e − 7 | 80.19 | 83.69 | 2.98 |
| 0 | 79.80 | 82.19 | 4.02 |

### A.2 Hyperparameters.

We provide in Table A details of the hyperparameters for reproducing the results. For a fair comparison, we set the number of prompts $N$ and the number of tokens $M$ to 10 and 8, respectively, and prepend prompts into the first 5 MSA blocks of ViT [2] for all settings, following VQ-Prompt [7]. To set the number of sampled prompts and balancing parameter, we perform a grid search over $N_s \in \{1, 5, 10, 20, 30, 40\}$ and $\lambda \in \{5e-5, 1e-5, 5e-6, 1e-6, 5e-7, 1e-7\}$, respectively. Note that we perform the grid search on a *separate cross-validation split* for 10 tasks on ImageNet-R [4], that is, we do not use any samples from the validation split. Based on the grid search, we set $\lambda$ to $1e-6$ and $N_s$ to 30 for all settings, including CIFAR-100 [9] and CUB-200 [15].

## B More results

### B.1 Hyperparameter analysis.

We provide in Tables B-D an analysis of each hyperparameter. We can see from Tables B and C that the performance is consistently strong for $M > 1$ and $N > 1$, whereas very small values of $M$ and $N$ (*e.g.*, 1) perform poorly due to the limited capacity of prompts. Table D shows that large values of $\lambda$ penalize the changes in the prompts overly,

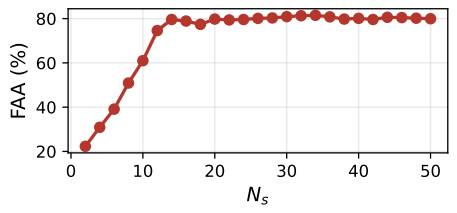


**Fig. A:** FAA w.r.t. $N_s$.

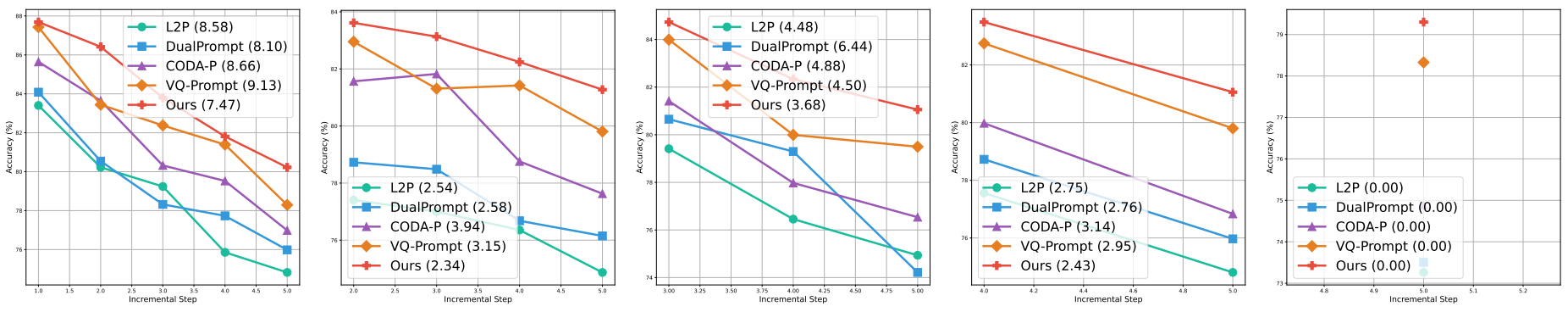


Task 1. Task 2. Task 3. Task 4. Task 5.

**Fig. B:** Per-task accuracy over incremental steps for the 5-task split on ImageNet-R [4]. Numbers in parentheses in the legend indicate the accuracy drop for each method, *i.e.*, the difference between the accuracies at the initial and final incremental steps of each task, where lower values indicate better preservation of previously learned knowledge.

**Table E:** Quantitative results for the 10 tasks setting in terms of FAA and $F$ on ImageNet-R [4]. All numbers are averaged over five runs with standard deviations.

| Method | FAA (↑) | $F$ (↓) |
|---|---|---|
| Joint-Training | 91.38 | - |
| L2P++ [18] | 72.67 ± 1.29 | 4.61 ± 0.85 |
| DualPrompt [17] | 75.33 ± 0.86 | 5.65 ± 0.53 |
| CODA-P [14] | 75.78 ± 0.68 | 6.33 ± 0.44 |
| VQ-Prompt [7] | 78.74 ± 0.26 | 4.29 ± 0.25 |
| **Ours** | **80.23** ± 0.31 | **3.09** ± 0.24 |

degrading performance. We further report the forgetting ratio $F$ in Table D, where the distribution regularization loss consistently reduces forgetting while maintaining overall accuracy. This suggests that our method better balances stability and plasticity by preventing abrupt changes in prompt distributions, rather than restricting plasticity. To further analyze the effect of the number of sampled prompts $N_s$, we provide in Fig. A experimental results covering a wider range of $N_s$. The results show that performance becomes stable once $N_s \geq 12$, while degrading in the low-$N_s$ regime. This is mainly because small values of $N_s$ inherently yield high-variance prompt estimates under stochastic sampling, leading to noisy optimization. We thus sample multiple prompts, *e.g.*, $N_s = 30$, to reduce the variance with negligible computational and memory overhead. Overall, our method is robust to the hyperparameters across a wide range of values, suggesting that it does not require extensive hyperparameter search.

### B.2 Forgetting ratio.

We compare in Table E our method with other prompt-based methods for a 10-task split on ImageNet-R [4]. We report the average forgetting ratio ($F$) in Eq. (iii) over five runs with standard deviations. We reproduce the numbers in the table by ourselves for a fair comparison. From the table, we can see that ours outperforms

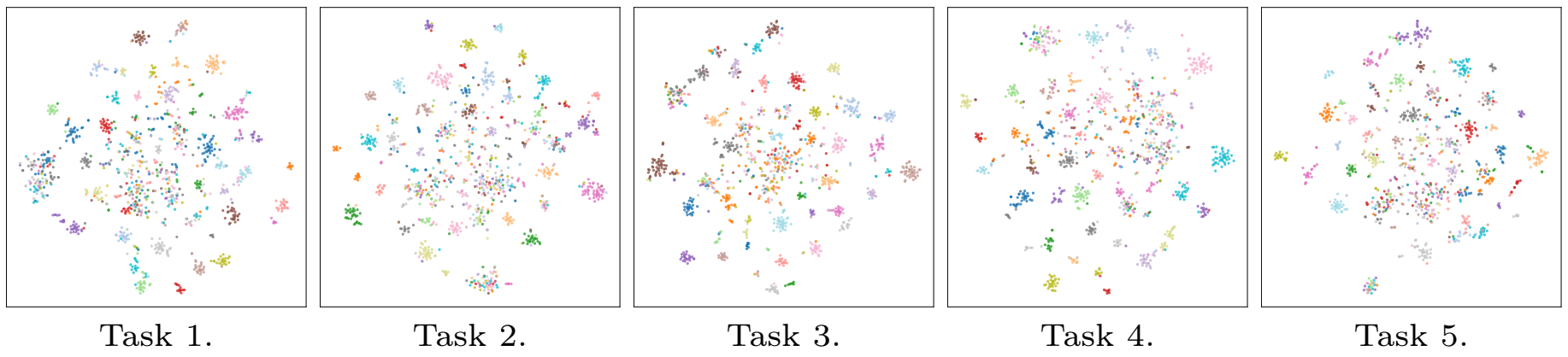


**Fig. C:** t-SNE visualization of query features obtained from each task for 5 tasks on ImageNet-R [4] (classes are color-coded). Each plot corresponds to one of the 5 tasks, indicating that images are highly diverse in the continual learning scenarios. Note that we obtain query features using ViT-B/16 [2] pre-trained on ImageNet-1K [1]. Best viewed by zooming in with color.

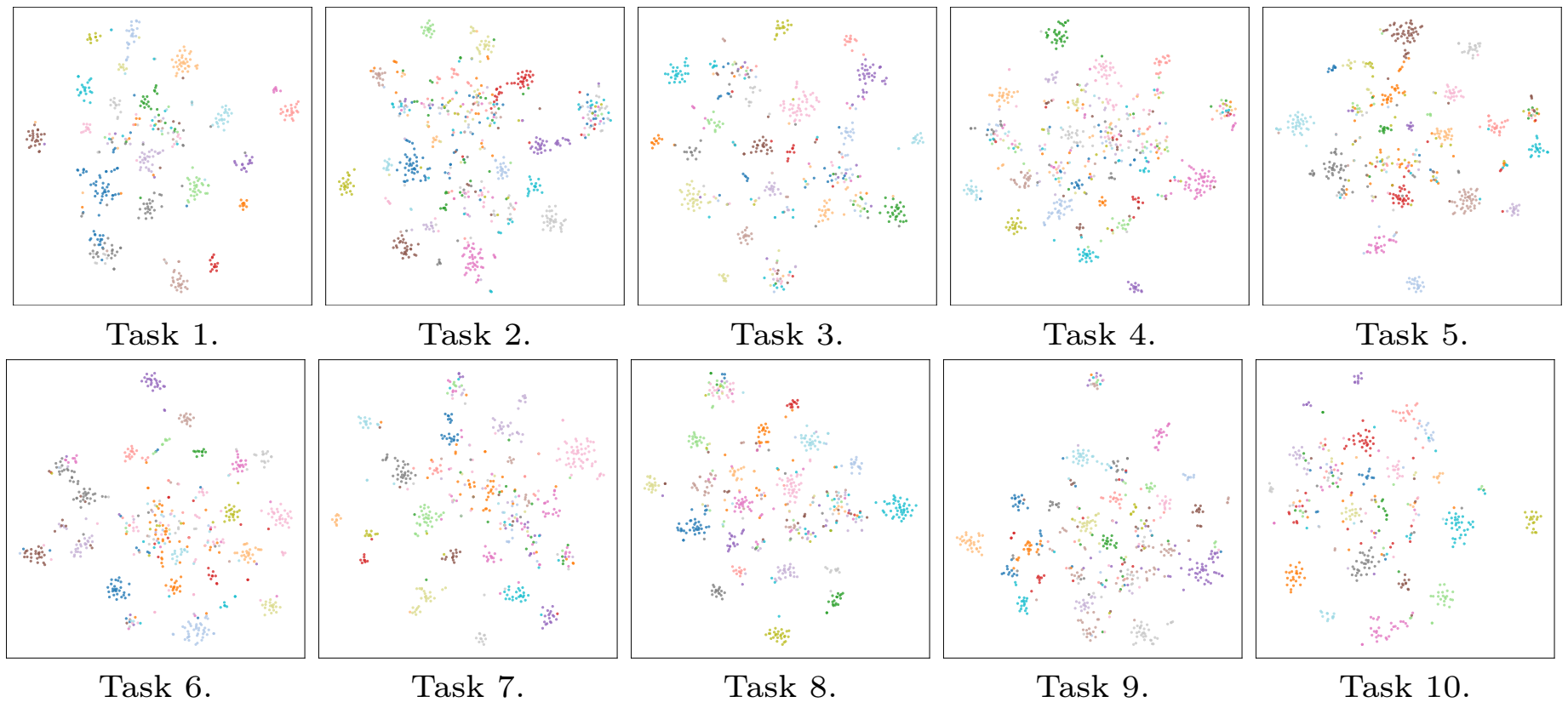


**Fig. D:** t-SNE visualization of query features, where we obtain them from each task for 10 tasks on ImageNet-R [4] (classes are color-coded). Each subplot indicates one of the 10 tasks. We can see that the diverse patterns are present in a sequence of data. We obtain query features using ViT-B/16 [2] pre-trained on ImageNet-1K [1]. Best viewed by zooming in with color.

the state-of-the-art prompt-based methods by clear margins in terms of FAA and the forgetting ratio $F$. This demonstrates that the distribution regularization loss mitigates abrupt changes in the learned distributions effectively, encouraging the distribution to retain previous knowledge. This allows our method to avoid the forgetting problem.

### B.3 Additional results on per-task accuracy over incremental steps.

We show in Fig. B per-task accuracy for the 5-task split on ImageNet-R [4]. The results show similar trends to those of the 10-task split (Fig. 4 in the main paper), where our method achieves higher accuracy across most tasks over the incremental steps. Furthermore, the accuracy drop in our method is consistently smaller than that of other methods, confirming that the distribution regularization loss effectively mitigates the forgetting problem across different task splits.

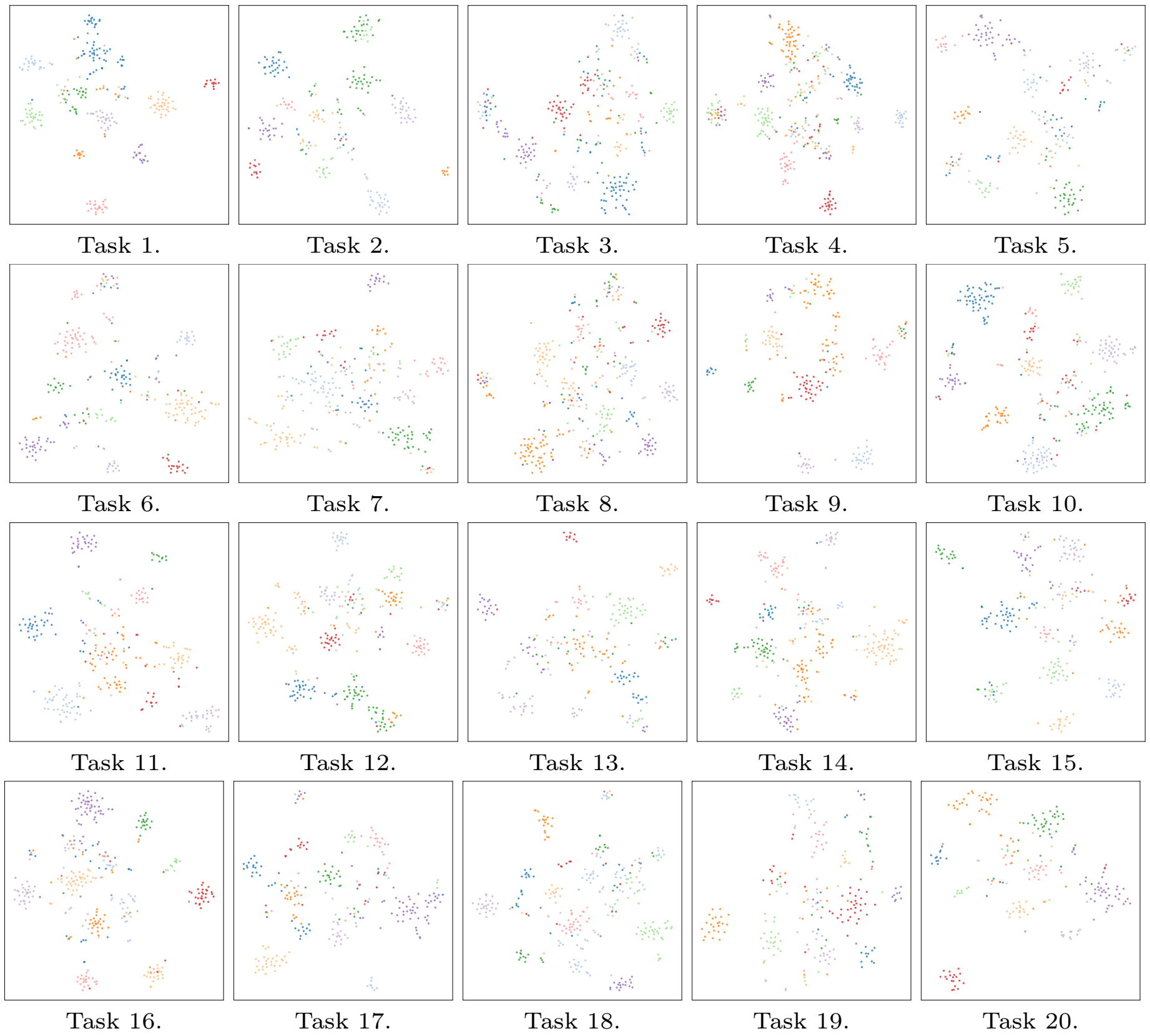

**Fig. E:** t-SNE visualization of query features obtained from each task for 20 tasks on ImageNet-R [4] (classes are color-coded). Each subplot shows the corresponding tasks, indicating the diverse data distributions in the continual learning process. We obtain query features using ViT-B/16 [2] pre-trained on ImageNet-1K [1]. Best viewed by zooming in with color.

# C More discussions

## C.1 Diverse patterns of images in the continual learning process.

In continual learning scenarios, a single task inherently exhibits high intra-task variations due to the presence of multiple distinct object classes. The feature variations between images become even more apparent as new tasks are sequentially introduced throughout the continual learning process, leading to increasingly complex data distributions that are difficult to capture (see Figs. C-E). This highlights the importance of capturing the diverse data distributions across tasks for prompt-based continual learning scenarios.

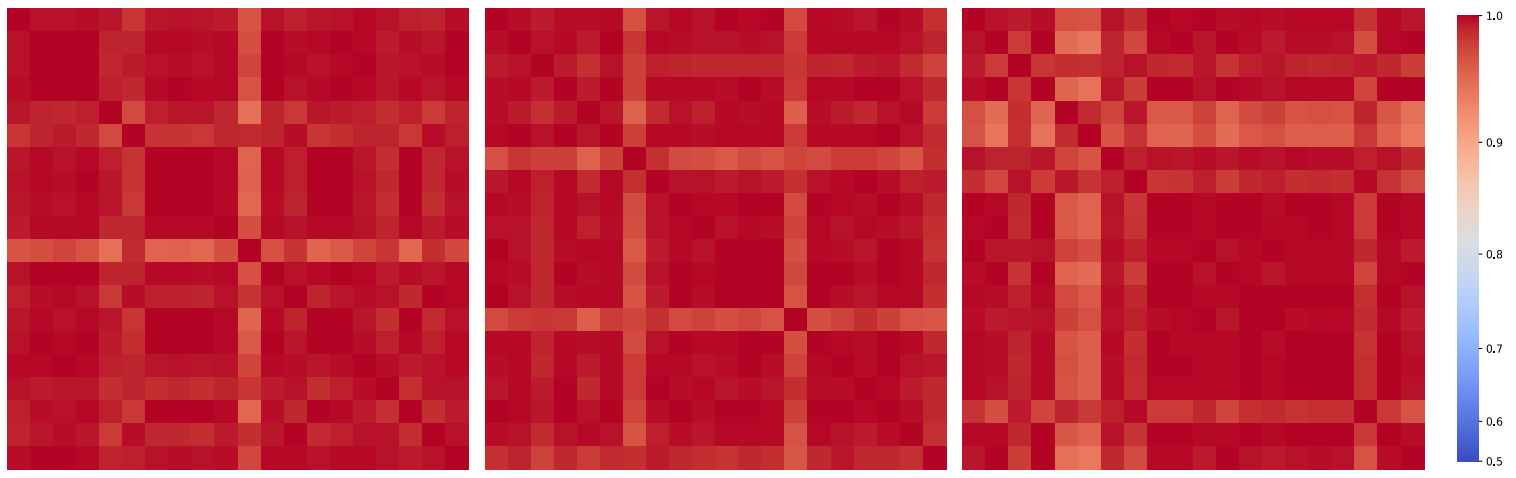

**(a)** CODA-P ($N = 100$). **(b)** CODA-P ($N = 150$). **(c)** CODA-P ($N = 200$).

**Fig. F:** Visualization of cosine similarity between prompts, obtained from CODA-P [14] using **(a)** 100, **(b)** 150, and **(c)** 200 component vectors, respectively. The prompts in **(a)**-**(c)** and Fig. 2 in the main paper are obtained for the same query features, where each query feature is randomly selected from different classes for 10 tasks of ImageNet-R [4].

### C.2 Empirical observation on prompt collapse.

The key factor for prompt-based continual learning is to capture the diverse data distributions across tasks by leveraging learned prompts (See Fig. C-E). However, existing prompt-based methods [7, 14, 17, 18] suffer from the *prompt collapse* problem, where prompts become highly similar to one another, despite diverse image patterns. From Fig. 2 in the main paper, we have observed that the prompts of the previous methods [7, 14, 17, 18] are highly similar to each other, with pairwise similarities of 1.0 for almost all pairs across all tasks. This indicates that the prompts of the existing methods are almost identical across the distributions of input queries, confirming the prompt collapse problem. They thus fail to capture the diverse data distributions across the sequence of tasks, leading to sub-optimal performance in continual learning (See Sec. 4 in the main paper). One might expect that increasing the number of component vectors (*e.g.*, $N$) could mitigate this issue, but the prompts remain highly similar even with more components (Fig. F), suggesting that prompt collapse cannot be addressed by simply scaling up the number of components $N$. By contrast, our method samples prompts from learned prompt distributions, resulting in relatively lower pairwise prompt similarities, effectively addressing the prompt collapse problem and enabling the model to capture the diverse data distributions across tasks.

### C.3 Effect of query-relevant weighting.

Our framework constructs a mixture of prompt distributions using the relevance scores in Eq. (2), which assign larger weights to distributions closer to a query feature. To examine whether the learned distributions capture meaningful information for prediction, rather than merely being diverse, we compare our relevance-based weighting with two variants on ImageNet-R [4]: uniform weighting,

**Table F:** Comparison of FAA w.r.t. the weighting schemes on ImageNet-R [4].

| Weighting | FAA |
|---|---|
| original | 80.35 |
| uniform | 66.99 |
| reversed | 11.04 |

which assigns an equal weight to all distributions, and reversed weighting, which assigns larger weights to distributions farther from the query feature. Both variants are applied at inference time to the same trained model, while keeping the training procedure unchanged. As shown in Table F, our method consistently outperforms both variants, and the performance degrades substantially as the weighting deviates further from the relevance-based scheme, *i.e.*, from uniform to reversed. This indicates that the learned distributions capture meaningful information for prediction, rather than merely being diverse.

### C.4 Comparison with probabilistic prompt modeling.

As discussed in Sec. 2, probabilistic prompt methods such as ProDA [11] and PPL [10] model the prompts themselves as deterministic vectors, while defining probabilistic distributions over their output representations. To compare such output-level modeling with our prompt-level modeling, we provide in Table G the results of three variants for the 10-task on ImageNet-R [4]: a deterministic baseline, an output-level variant, and our prompt-level modeling. The output-level variant applies probabilistic modeling to the output representations induced by the prompts, similar to ProDA [11] and PPL [10], while keeping the prompts deterministic. The other components remain identical to our method, so that the three variants differ only in where the probabilistic modeling is applied. We also show in Table G the pairwise prompt similarity, which quantifies the prompt collapse observed in Fig. 2. From the table, we can see that the output-level variant still suffers from high prompt similarity, whereas ours reduces it by clear margins, leading to better performance. This indicates that modeling the prompts themselves as distributions is crucial for addressing prompt collapse, which arises at the prompt level.

**Table G:** Comparison of prompt-level and output-level modeling on ImageNet-R [4].

| Method | FAA | Prompt similarity |
|---|---|---|
| Deterministic | 77.58 | 0.99 |
| Output-level | 78.45 | 0.97 |
| Prompt-level | 80.35 | 0.83 |

We further examine whether the gains of our framework come from its distributional formulation, rather than simply from adding stochasticity or diversity. To this end, we provide in Table H the results of deterministic baselines augmented with Gaussian perturbation or an explicit prompt-diversity loss [11] on ImageNet-R [4]. From the table, we can clearly see that our method consistently outperforms both baselines. This suggests that neither injecting stochasticity through perturbation nor imposing an explicit diversity constraint is sufficient to address prompt collapse, and that the effectiveness of our framework instead stems from its query-conditioned distributional formulation.

**Table H:** Comparison with deterministic baselines augmented with perturbation or a diversity loss on ImageNet-R [4].

| Method | FAA |
|---|---|
| Deterministic | 77.58 |
| w/ perturb | 77.83 |
| w/ div. loss | 78.14 |
| Ours | 80.35 |

### C.5 Analysis of the aggregation scheme.

We provide here theoretical and empirical analyses of the proposed aggregation scheme in Eqs. (9) and (10) in the main paper.

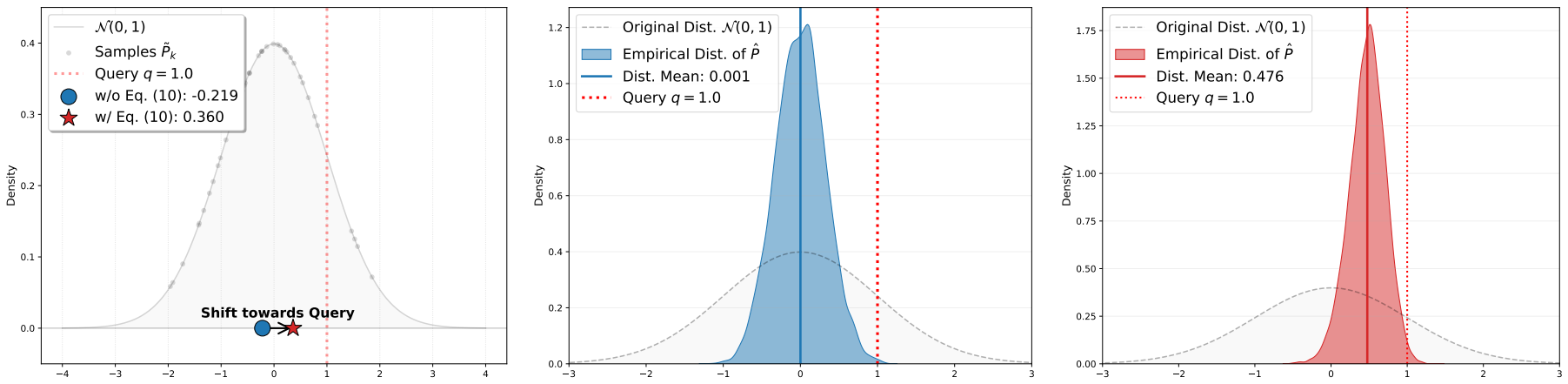


**(a)** Simple avg. vs. proposed. **(b)** Simple avg. ($w_k = 1/N_s$). **(c)** Proposed (Eq. (10)).

**Fig. G:** Visualization of the aggregation scheme using a 1D toy experiment, where prompts $\tilde{P}_k$ are sampled from $\mathcal{N}(0,1)$ with a fixed query $q = 1.0$. **(a)** Comparison between the simple average (blue circle) and the proposed aggregation (red star), where the prompt obtained from the proposed aggregation is closer to the query. Empirical distributions of $\hat{P}$ obtained by **(b)** the simple average and **(c)** the proposed aggregation over repeated experiments, centered around 0.001 and 0.476, respectively. We can see that the proposed aggregation is significantly closer to the query $q = 1.0$ compared to the simple average. Best viewed by zooming in with color.

**Theoretical analysis.** For a given $N_s$, the simple average $\hat{P} = \frac{1}{N_s}\sum_{k=1}^{N_s}\tilde{P}_k$ is an unbiased estimator of the mixture mean $\mu_{\mathrm{GM}}$:

$$\mathbb{E}\left[\hat{P}\right] = \mathbb{E}\left[\frac{1}{N_s}\sum_{k=1}^{N_s}\tilde{P}_k\right] = \mu_{\mathrm{GM}}. \tag{iv}$$

This indicates that the expected value of the simple average is determined solely by $\mu_{\mathrm{GM}}$, regardless of the query feature $q$. In contrast, the proposed aggregation assigns higher weights to samples that are more similar to $q$:

$$\mathbb{E}\left[\hat{P}\right] = \mathbb{E}\left[\sum_{k=1}^{N_s} w_k \tilde{P}_k\right] = \mathbb{E}_{\tilde{P}\sim\mathcal{N}_{\mathrm{GM}}}\left[\frac{\exp(\mathrm{sim}(\tilde{P},q))}{\mathbb{E}[\exp(\mathrm{sim}(\tilde{P},q))]}\tilde{P}\right], \tag{v}$$

which shifts the expected value of $\hat{P}$ towards the query feature $q$, compared to $\mu_{\mathrm{GM}}$. Thus, by drawing $N_s$ independent samples, our aggregation scheme reflects the characteristics of the query feature while preserving the diversity of the sampled prompts.

**Empirical analysis.** We verify the above analysis using a 1D toy experiment, where we sample $N_s = 30$ prompts from $\mathcal{N}(0,1)$ with a fixed query $q = 1.0$, and compare prompts obtained from the simple average and the proposed aggregation scheme. Fig. G(a) illustrates a comparison between the two aggregation schemes, where the simple average yields $-0.219$ and the proposed aggregation yields 0.360. We can see that the prompt obtained from the proposed aggregation is significantly closer to the query $q = 1.0$, with a distance of 0.640, compared to a distance of 1.219 for the simple average. This demonstrates that the proposed aggregation scheme effectively shifts the final prompt towards the query feature, whereas the

simple average fails to reflect the query due to the inherent randomness of the sampling process. To further analyze the behavior of each aggregation scheme, we repeat the experiment 10,000 times and show the empirical distributions of $\hat{P}$ in Figs. G(b) and (c). The simple average is centered around 0.001, close to the mean of $\mathcal{N}(0,1)$ and far from the query, whereas the proposed aggregation is centered around 0.476, significantly closer to the query. These results confirm that the proposed aggregation scheme consistently produces prompts that better reflect the characteristics of the query feature than those of the simple average, while preserving the diversity of the sampled prompts by retaining contributions from all samples.

### C.6 Moment matching.

Approximating a Gaussian mixture with a single Gaussian via moment matching is a standard practice in probabilistic modeling [3, 5, 12, 13], as it retains the first and second-order statistics of the mixture, while allowing all components to be jointly optimized during training. Following this convention, we define the mixture distribution in Eqs. (4)-(7) in the main paper by matching the first and second moments of the Gaussian mixture using the law of total variance as follows:

$$\mu_{\text{GM}} = \mathbb{E}_{\text{GM}}[x] = \sum_{n=1}^{N} s(n)\, \mathbb{E}_{\mathcal{N}(\mu(n),\Sigma(n))}[x] = \sum_{n=1}^{N} s(n)\, \mu(n), \tag{vi}$$

and

$$\begin{aligned} \sigma^2_{\text{GM}} &= \text{Var}_{\text{GM}}[x] \\ &= \underbrace{\sum_{n=1}^{N} s(n)\, \sigma^2(n)}_{\text{within-component variance}} + \underbrace{\sum_{n=1}^{N} s(n)\, (\mu(n) - \mu_{\text{GM}})^2}_{\text{between-component variance}} \\ &= \sum_{n=1}^{N} s(n) \left(\sigma^2(n) + (\mu(n) - \mu_{\text{GM}})^2\right), \end{aligned} \tag{vii}$$

where the variance decomposes into within-component and between-component contributions, respectively, and $x$ is a random variable.

To further support this practice empirically, we compare in Table I moment matching and GMM, where we implement the GMM baseline using the Gumbel-Softmax trick [6] with temperature annealing, where a temperature is exponentially decayed from 1.0 to 0.1 over training. From the table, we can see that moment matching achieves comparable or slightly better performance, confirming that the approximation incurs no performance degradation in our setting.

**Table I:** Comparison of FAA and CAA w.r.t. mixture. GMM: Gaussian mixture model with Gumbel-Softmax. MM: Moment matching.

| Sampling from | FAA (↑) | CAA (↑) |
|---|---|---|
| GMM | 80.17 | 83.65 |
| MM | **80.35** | **83.88** |

**Algorithm 1** Sampling

**Input**: Sequence of tasks, Pre-trained backbone, $M \times N$ prompt distributions $\mathcal{N}(\mu(m,n), \Sigma(m,n))$, $K$ for top-$K$ selection, and number of sampled prompts $N_s$.

**for** $t = 1$ to $T$ **do**
  **for** $i = 1$ to $I$ **do**
    Extract query feature $q$ from an input image using the backbone.
    **for** $m = 1$ to $M$ **do**
      Compute relevance score $s(n)$ for the $n$-th prompt distribution w.r.t. $q$ as in Eq. (2).
      Form a mixture of distributions $\mathcal{N}_{\text{GM}}$ as in Eq. (4), using Eqs. (5)–(7).
      Sample $N_s$ probabilistic prompts based on the reparameterization trick [8] as in Eq. (8).
      Aggregate the $N_s$ sampled prompts using Eqs. (9)–(10).
    **end for**
    **if** training **then**
      Compute the regularization term as in Eqs. (13)–(14).
      Cache the current distribution and then update each prompt distribution.
    **end if**
    Prepend the aggregated prompt in Eq. (11) using the prefix tuning technique, concatenate the prompt with input tokens, and input them into the pre-trained model.
  **end for**
**end for**

### C.7 Sampling process.

We provide in Algorithm 1 the overall sampling process of our framework. Given a query feature $q$ from an input image, we form a mixture of prompt distributions to sample prompts for each token. Specifically, we first compute the relevance score $s(n)$ for the $n$-th prompt of each token, based on the distance between the query feature and the prompt distributions. We then form a mixture distribution for each token using the scores. From the mixture distribution, we sample $N_s$ probabilistic prompts for each token using the reparameterization trick [8]. This helps to improve learning stability by reducing the variance induced by the stochastic sampling process. We aggregate the sampled prompts considering the relation between the samples and the query to preserve the diversity of samples while reflecting the characteristics of the query feature. Finally, we prepend the aggregated sampled prompts with the input query, and optimize them using the Pre-T prompting technique. Note that the sampling process incurs no significant memory and computational costs, as we aggregate the sampled prompts before prompting, and thus the sampling process requires only a single forward pass (Sec. 4.3 in the main paper).



\end{document}